\documentclass{article}

\usepackage{microtype}
\usepackage{graphicx}
\usepackage{subcaption}
\usepackage{booktabs} % for professional tables
\usepackage{hyperref}
\usepackage{color, soul}
\usepackage{kotex}
\usepackage{subcaption}

\usepackage{algorithmic}

\renewcommand{\algorithmiccomment}[1]{\hfill$\triangleright$~#1}

\usepackage[preprint]{icml2026}

\usepackage{adjustbox}
\usepackage{amsmath}
\usepackage{amssymb}
\usepackage{mathtools}
\usepackage{amsthm}
\usepackage{booktabs}
\usepackage{makecell}
\usepackage{multirow}
\usepackage{placeins}
\usepackage{fvextra} 
\usepackage{tabularx}
\usepackage{array}
\usepackage{ragged2e}
\DefineVerbatimEnvironment{promptblock}{Verbatim}{
  fontsize=\small,
  breaklines=true,
  breakanywhere=true
}

\usepackage[capitalize,noabbrev]{cleveref}

\usepackage{xcolor}
\usepackage[most]{tcolorbox}
\tcbuselibrary{listings,breakable}

\newtcblisting{mycodebox}{
  colback=white!95!black,
  left=0pt,right=0pt,
  listing only,
  listing engine=listings,
  breakable,
  listing options={
    basicstyle=\ttfamily\scriptsize,
    columns=fullflexible,
    keepspaces=true,
    breaklines=true,
    breakindent=8pt,
    breakautoindent=false,
    escapeinside={(*@}{@*)}
  }
}

\newtcolorbox{benchbox}{
  enhanced,
  breakable,
  colback=white,
  colframe=black,
  boxrule=1pt,
  arc=3mm,
  left=6pt, right=6pt, top=6pt, bottom=6pt,
  boxsep=2pt
}

\theoremstyle{plain}

\theoremstyle{definition}

\theoremstyle{remark}

\usepackage[textsize=tiny]{todonotes}

\icmltitlerunning{Models Know Models Best: Evaluation via Model-Preferred Formats}

\begin{document}

\twocolumn[
  % \icmltitle{Models Know Models Best: Leveraging LLMs for LLM Evaluation}
  \icmltitle{Models Know Models Best: Evaluation via Model-Preferred Formats}


  \icmlsetsymbol{equal}{*}

  \begin{icmlauthorlist}
    \icmlauthor{Joonhak Lee}{equal,ds}
    \icmlauthor{Sungmok Jung}{equal,ds}
    \icmlauthor{Jongyeon Park}{equal,ds}
    \icmlauthor{Jaejin Lee}{ds,cs}
  \end{icmlauthorlist}

  \icmlaffiliation{ds}{Graduate School of Data Science, Seoul National University}
  \icmlaffiliation{cs}{Dept. of Computer Science and Engineering, Seoul National University}

  \icmlcorrespondingauthor{Joonhak Lee}{hmjelee@snu.ac.kr}
  \icmlcorrespondingauthor{Sungmok Jung}{tjdahrwjd@snu.ac.kr}
  \icmlcorrespondingauthor{Jongyeon Park}{iscopark67@snu.ac.kr}
  \icmlcorrespondingauthor{Jaejin Lee}{jaejin@snu.ac.kr}
  \icmlkeywords{Machine Learning, ICML}

  \vskip 0.3in
]



\printAffiliationsAndNotice{\icmlEqualContribution}

\begin{abstract}
Performance of Large Language Models (LLMs) on multiple-choice tasks differs markedly between symbol-based and cloze-style evaluation formats. The observed discrepancies are systematically attributable to task characteristics: natural language continuation benefits from likelihood scoring, whereas explicit comparison is better suited to symbol-based selection. These trends are consistent across various decoder-based LLMs, indicating model-agnostic effects. To address these inconsistencies, a dynamic format-alignment strategy is introduced that employs a lightweight classifier trained on latent model-preference signals. In contrast to human-designed heuristics, which often degrade performance, this approach uses model-generated signals to determine the optimal format for each problem instance. The proposed method achieves substantial and consistent improvements in zero-shot accuracy across reasoning and knowledge benchmarks, better revealing the models' latent capabilities.

\end{abstract}
\section{Introduction}
Multiple-choice question answering (MCQA) is one of the most widely used methods for evaluating and comparing large language models (LLMs). Its clear evaluation protocol and alignment with human problem-solving processes have contributed to its status as a de facto standard in LLM evaluation~\citep{lyu2024beyond}.

However, an increasing number of studies raise concerns about whether MCQA accurately reflects LLMs' true capabilities. Researches have shown that model performance can be highly sensitive to small changes in input~\cite{sclarquantifying,zheng2023large,pezeshkpour2024large}, and that models can attain misleadingly high accuracy even when the question is partially or entirely removed~\cite{belinkov2019don,feng2019misleading,srikanth2022partial,cho2026choicesspeaklouderquestions}. These findings cast doubt on the reliability of MCQA as an accurate measure of reasoning ability.

In addition to these concerns, the MCQA framework accepts various evaluation formats, a relatively underexplored aspect. Typical formats include a symbol-based approach, in which the model selects from lettered options (e.g., A, B, C, and D), and a cloze-style approach, in which evaluation is based on the model's likelihood assigned directly to the answer options. Recent studies have highlighted significant performance gaps between these two formats~\cite{robinson2023leveraging,gu2025olmes}, yet the reasons behind these discrepancies and their patterns are not well understood.

In this study, we conduct a thorough empirical comparison of symbol-based and cloze-style MCQA methods across a variety of models and benchmarks. Our findings indicate that the performance differences, which depend on how the task is formulated, are closely linked to the characteristics of the task. As a result, we observe stable and consistent preferences for certain formats across different models. Specifically, tasks that require completing natural language continuations show significant performance drops when using symbol-based formats compared to cloze-style ones. Although providing demonstration examples can help reduce these discrepancies, they do not completely eliminate the challenges that models encounter when evaluation methods are not aligned with the nature of the task.

Driven by these insights, we explore whether choosing the evaluation format at the problem instance level can enhance model performance. We consider two approaches: one that employs human-designed linguistic heuristics and the other that derives format preferences directly from model behavior. Our findings reveal that human-aligned linguistic criteria do not consistently lead to improvements, whereas a classifier trained on signals of model preferences achieves significant gains. These improvements are particularly noticeable in completion-oriented tasks, where format mismatches are most pronounced.

Finally, we demonstrate that preferences for various formats, inferred from a subset of language models, can be generalized to unseen models through majority voting. This indicates that decoder-based language models exhibit task-dependent biases in their evaluation formats, rather than unique, model-specific behaviors. By using these shared preferences, we can achieve a more accurate and insightful assessment of language model capabilities and emphasize the importance of aligning evaluation formats with the inductive biases of contemporary language models.

The contributions of this paper are summarized as follows: 
\begin{itemize} 
\setlength{\itemsep}{0pt} \setlength{\parskip}{0pt} 
\item We establish that performance discrepancies arising from format mismatch in MCQA are not stochastic but vary systematically according to the structural and linguistic properties of the task. We demonstrate that these performance patterns are model-agnostic, emerging consistently across a diverse suite of decoder-based LLMs.
\item We provide evidence that the task characteristics governing format preference are implicitly encoded within the model's latent representations. We show that by explicitly eliciting these latent preferences, it is possible to recover significant performance otherwise lost to suboptimal formatting.
\item We propose a format-alignment strategy that leverages these latent signals to dynamically select the optimal evaluation format. The proposed approach yields substantial and consistent gains in zero-shot accuracy across a broad spectrum of reasoning and knowledge-intensive benchmarks, effectively bridging the gap between model capability and task characteristics.
\end{itemize}

\section{Related Work}

\subsection{MCQA Evaluation Formats: \emph{Symbol} and \emph{Cloze}}

MCQA has become a widely accepted standard evaluation framework for LLMs~\cite{achiam2023gpt,team2023gemini,jiang2023mistral7b,grattafiori2024llama3herdmodels}. In the MCQA formats, model outputs are limited to a predefined set of answer options, and predictions are assessed by comparing the likelihoods assigned to each option. Compared to open-ended generation, this constrained setup provides a precise, reproducible evaluation criterion, which contributes to the widespread adoption of MCQA for benchmarking modern LLMs~\citep{lyu2024beyond}.

MCQA is typically presented in two formats: symbol-based and cloze-style~\cite{alzahrani2024benchmarks}. In the symbol-based format, the prompt includes a question followed by a list of candidate answers. The model selects an option by generating the symbol (e.g., A, B, C, or D) with the highest likelihood. On the other hand, the cloze-style format treats the question as a prompt and evaluates each answer option by calculating its likelihood when appended to the prompt. The option with the highest likelihood is chosen as the model's prediction. Because answer options in the cloze format often vary in length, scores are usually normalized based on the number of characters~\cite{touvron2023llama,sutawika2025eleutherai} or tokens~\cite{brown2020language}  to reduce any biases related to the length.

\subsection{Sensitivity to MCQA Evaluation Formats}
Although it may seem straightforward, MCQA is highly sensitive to formatting decisions. Previous studies show that model performance can differ significantly with minor changes, such as rephrasing prompts~\cite{sclarquantifying}, altering the order of answer options~\cite{zheng2023large,pezeshkpour2024large}, or modifying a few-shot demonstration~\cite{lu2022fantastically}. To address this inconsistency, several strategies have been proposed to aggregate predictions across different permutations during inference, such as using majority voting~\cite{zong2024fool,wei2024unveiling,zhou2024large}. Additionally, methods to enhance robustness during training include techniques like knowledge distillation~\cite{liusie2024teacher} and structural pruning~\cite{choi2025mitigating}.

More recent studies have shown that a single model can exhibit significantly different performance on the same MCQA task, depending on whether a symbol-based or cloze-style format is used. These differences prompt important questions about whether models are truly reasoning about the task content, especially since identical underlying questions can yield different outcomes simply due to changes in the evaluation format.

Existing studies provide limited insight into the mechanisms driving performance gaps across different formats. Simply using the format that performs best for a specific benchmark, as noted by ~\citet{gu2025olmes}, only reports the phenomenon without explaining it. Likewise, attributing the effectiveness of symbol-based formats to enhanced symbol-option binding capabilities, as discussed by ~\citet{robinson2023leveraging}, does not adequately address situations where cloze-style formats consistently outperform symbol-based ones, even in models with strong binding abilities. 

In contrast to previous studies, we conduct a systematic analysis across multiple models and benchmarks to understand the structure of performance differences that depend on evaluation formats. We also demonstrate how these insights can be used to enhance MCQA evaluation by selecting the evaluation format in a principled and instance-aware manner.

\section{Symbol vs. Cloze: A Baseline Study}
In this section, we conduct a systematic empirical comparison of symbol-based evaluation and cloze-style evaluation across various tasks and model families. 
%By assessing each model using both evaluation methods, we analyze how the measured performance varies based on the evaluation format and whether consistent trends emerge at the task level.

\subsection{Evaluation Setup}
\paragraph{Models.} To investigate how task-dependent differences in evaluation formats vary across different models, we assess a total of 22 language models. Our experiments include several variants from the Qwen 3~\cite{yang2025qwen3}, Llama 3.1 and 3.2~\cite{grattafiori2024llama3herdmodels}, and Mistral~\cite{jiang2023mistral7b} model families. For more details about the evaluated models, please refer to Appendix~\ref{app:models}.

\begin{table*}[!t]
\centering
\resizebox{\textwidth}{!}{%
\begin{tabular}{llrrr}
\toprule
\textbf{Name} & \textbf{Description} & \textbf{Train} & \textbf{Validation} & \textbf{Test} \\
\midrule
ARC-Easy\cite{clark2018think} & Multiple-choice science questions requiring factual retrieval or logical reasoning & 2,251  & 570  & 2,376 \\ 
ARC-Challenge\cite{clark2018think} & Multiple-choice science questions requiring factual retrieval or logical reasoning & 1,119  & 299  & 1,172 \\ 
MMLU\cite{hendrycks2021measuring} & Broad-scale multiple-choice evaluation covering 57 academic and professional subjects
  & 99,842 & 1,531 & 14,042 \\ 
OpenBookQA\cite{mihaylov2018can} & Multiple-choice science tasks requiring integration of external facts with reasoning
  & 4,957  & 500  & 500 \\ 
CommonsenseQA\cite{talmor2019commonsenseqa} & Multiple-choice reasoning based on semantic knowledge of everyday objects
& 7,792  & 1,948 & 1,221 \\ 
HellaSwag\cite{zellers2019hellaswag} & Continuation task requiring models to predict the most plausible narrative ending
  & 31,924 & 7,981 & 10,042 \\ 
PIQA\cite{bisk2020piqa} &Continuation task focused on physical interactions and goal-oriented completion & 12,890 & 3,222 & 1,838 \\
WinoGrande\cite{sakaguchi2021winogrande} & Continuation task using fill-in-the-blank pairs to resolve coreference ambiguity
  & 32,318 & 8,079 & 1,267 \\ 
\bottomrule
\end{tabular}
}
\caption{Benchmarks used in the evaluation, including task descriptions and dataset split statistics.}
\label{tab:benchmarks}
\end{table*}

\paragraph{Benchmarks.} 
We evaluate model performance using eight MCQA benchmarks that are commonly used to assess LLMs. These datasets encompass a wide variety of domains, including scientific reasoning, general knowledge, professional-level understanding, and different types of commonsense reasoning. By covering diverse reasoning tasks and knowledge areas, our evaluation aims to capture a broad range of characteristics relevant to MCQA performance. To ensure reliable and reusable evaluation results, we clearly differentiate between training, validation, and test splits for all tasks. When a dataset includes a labeled test split, we evaluate the models on that original test set. If a labeled test set is not provided, we treat the original validation split as the test set and further divide the original training split into new training and validation sets using an 80:20 ratio. For detailed descriptions and split statistics, please refer to Table~\ref{tab:benchmarks}.

\begin{figure}[!t]
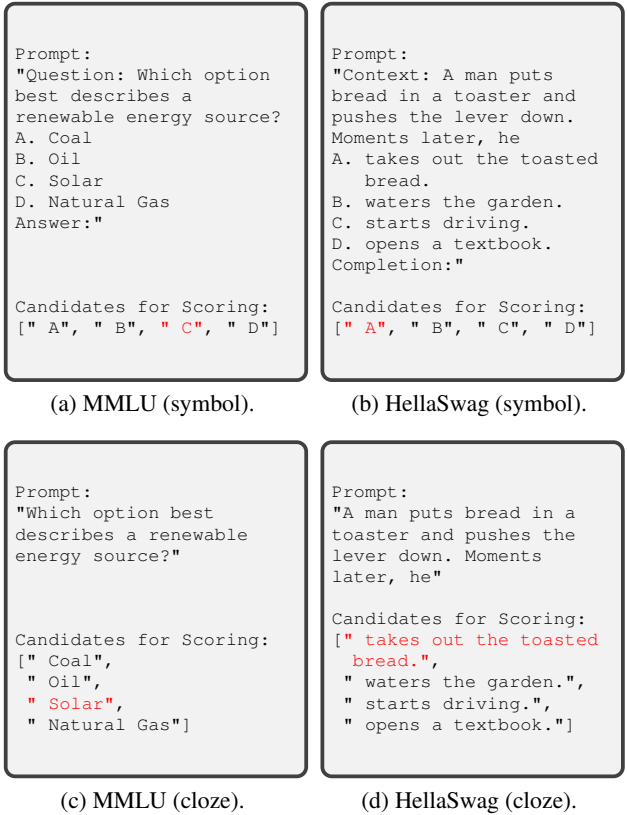

\centering
\begin{scriptsize}
\begin{minipage}[t]{\linewidth}
\centering

\begin{minipage}[b]{0.49\linewidth}
\centering
\begin{mycodebox}
Prompt:
(*@\textcolor{black}{"}@*)Question: Which option 
best describes a 
renewable energy source?
A. Coal
B. Oil
C. Solar
D. Natural Gas
Answer:(*@\textcolor{black}{"}@*)



Candidates for Scoring:
[(*@\textcolor{black}{" A"}@*), (*@\textcolor{black}{" B"}@*), (*@\textcolor{red}{" C"}@*), (*@\textcolor{black}{" D"}@*)]
\end{mycodebox}
\begin{small}
(a) MMLU (symbol).
\end{small}
\end{minipage}
\hfill
\begin{minipage}[b]{0.49\linewidth}
\centering
\begin{mycodebox}
Prompt:
(*@\textcolor{black}{"}@*)Context: A man puts 
bread in a toaster and 
pushes the lever down. 
Moments later, he
A. takes out the toasted 
   bread.
B. waters the garden.
C. starts driving.
D. opens a textbook.
Completion:(*@\textcolor{black}{"}@*)

Candidates for Scoring:
[(*@\textcolor{red}{" A"}@*), (*@\textcolor{black}{" B"}@*), (*@\textcolor{black}{" C"}@*), (*@\textcolor{black}{" D"}@*)]
\end{mycodebox}
\begin{small}
(b) HellaSwag (symbol).
\end{small}
\end{minipage}

\vspace{\baselineskip}

\begin{minipage}[b]{0.49\linewidth}
\centering
\begin{mycodebox}
Prompt:
(*@\textcolor{black}{"}@*)Which option best 
describes a renewable 
energy source?(*@\textcolor{black}{"}@*)



Candidates for Scoring:
[(*@\textcolor{black}{"}@*) Coal(*@\textcolor{black}{"}@*), 
 (*@\textcolor{black}{"}@*) Oil(*@\textcolor{black}{"}@*), 
 (*@\textcolor{red}{" Solar"}@*), 
 (*@\textcolor{black}{"}@*) Natural Gas(*@\textcolor{black}{"}@*)]
\end{mycodebox}
\begin{small}
(c) MMLU (cloze).
\end{small}
\end{minipage}
\hfill
\begin{minipage}[b]{0.49\linewidth}
\centering
\begin{mycodebox}
Prompt:
(*@\textcolor{black}{"}@*)A man puts bread in a 
toaster and pushes the 
lever down. Moments
later, he(*@\textcolor{black}{"}@*)

Candidates for Scoring:
[(*@\textcolor{red}{" takes out the toasted bread."}@*),
 (*@\textcolor{black}{"}@*) waters the garden.(*@\textcolor{black}{"}@*),
 (*@\textcolor{black}{"}@*) starts driving.(*@\textcolor{black}{"}@*),
 (*@\textcolor{black}{"}@*) opens a textbook.(*@\textcolor{black}{"}@*)] 
\end{mycodebox}
\begin{small}
(d) HellaSwag (cloze).
\end{small}
\end{minipage}

\end{minipage}
\end{scriptsize}
\caption{Examples of prompts and candidates used for log-likelihood scoring in MMLU and HellaSwag under \texttt{symbol} and \texttt{cloze} evaluation formats. Correct answers are highlighted in red.}
\label{fig:prompt_example}
\end{figure}

% What is the meaning of Evaluated Outputs? why they do not have choices in the cloze format?

We evaluate each benchmark using both symbol-based and cloze-style MCQA formats. 

\paragraph{Symbol-based evaluation format.}
In the symbol-based setting, we construct the prompt by appending labeled answer choices to the question. This approach reflects how humans typically tackle multiple-choice questions, with the model selecting an answer from the provided labels. We report performance using standard accuracy for this format.
% (denoted as \textit{acc})  

\paragraph{Cloze-style evaluation format.}
In the cloze-style setting, the question is presented as a prompt, and the model computes the log-likelihood of each answer choice independently, selecting the option with the highest score. To address biases introduced by answer length, we assess performance using length-normalized accuracy, calculated by dividing the raw log-likelihood of each option by the number of tokens~\cite{brown2020language}. 
% (denoted as \textit{acc\_norm})

\paragraph{Prompt construction.}
We define task-specific prefixes (e.g., \texttt{Question} and \texttt{Context}) and suffixes (e.g., \texttt{Answer} and \texttt{Completion}) in a principled manner based on each benchmark's characteristics. For example, in HellaSwag~\cite{zellers2019hellaswag}, which requires selecting an appropriate continuation given a context, we use \texttt{Context} as the prefix and \texttt{Completion} as the suffix. Figure~\ref{fig:prompt_example} provides illustrative examples of both formats, while complete prompt templates for all benchmarks are detailed in Appendix~\ref{app:prompts}.

\paragraph{Few-shot evaluation.} We further analyze how the impact of MCQA formats manifests in in-context learning, where language models learn to perform a task based on a small number of demonstration examples provided in the prompt~\cite{brown2020language}. Our evaluation includes both zero-shot and few-shot settings, considering configurations with 1, 2, 5, and 10 examples. For the few-shot evaluations, demonstration examples are randomly sampled from the validation split using five different seeds (308, 713, 777, 1,234, and 4,649), and we report the average performance across these seeds.

\subsection{Findings}
\label{subsec:3_2_findings}
Our evaluation results in Table~\ref{tab:3_0shot_sym_vs_clz} show significant performance differences between the two methods, with these differences varying by task. We analyze these discrepancies by considering the unique characteristics of each task, emphasizing how task characteristics interact with different evaluation approaches. Additionally, we explore how these differences change under in-context learning, investigating whether few-shot demonstrations enhance or lessen the effects that depend on the evaluation format. Detailed evaluation results can be found in Appendix~\ref{app:baseline}.

\paragraph{Tasks have evaluation format preferences.}
We find consistent task-specific differences across evaluation formats for nearly all the models we evaluated. Specifically, as shown in Table~\ref{tab:3_0shot_sym_vs_clz}, the tasks ARC, MMLU, OpenBookQA, and CommonsenseQA demonstrate significantly higher performance with the symbol-based format. In contrast, tasks like HellaSwag, PIQA, and WinoGrande consistently perform better with the cloze-style format. Tasks that benefit from the symbol-based format often feature a multiple-choice structure. For instance, questions that ask, "Which of the following is …?" or include options like "None of the above" require all answer choices to be presented together in the prompt for a meaningful evaluation.

On the other hand, tasks that prefer the cloze-style format typically involve completing a natural-language continuation given a context. The notably poor performance of symbol-based formats on tasks such as HellaSwag and PIQA highlights a significant misalignment between the evaluation format and the task's nature, particularly when models are asked to solve natural continuation problems in a symbol-based setup. We hypothesize that this discrepancy arises from the autoregressive training paradigm of LLMs, which are optimized to predict natural continuations from extensive text corpora. As a result, cloze-style evaluations align more closely with the models' learned behaviors, resulting in consistently higher performance on these tasks.

% tables/task_means_0shot_symbol_cloze.tex
\begin{table}[!t]
\centering
\small
\resizebox{0.9\linewidth}{!}{%
\begin{tabular}{lrrr}
\toprule
\textbf{Task} & \textbf{Symbol (\%)} & \textbf{Cloze (\%)} & \textbf{$\Delta$ (\%p)} \\
\midrule
\multicolumn{4}{l}{\emph{Symbol-favoring tasks}} \\
ARC-Easy        & 89.1 & 62.6 & $+26.5$ \\
ARC-Challenge   & 77.8 & 41.5 & $+36.3$ \\
MMLU            & 63.6 & 40.1 & $+23.5$ \\
OpenBookQA      & 73.5 & 43.9 & $+29.6$ \\
CommonsenseQA   & 69.4 & 39.7 & $+29.7$ \\
\midrule
\multicolumn{4}{l}{\emph{Cloze-favoring tasks}} \\
HellaSwag       & 34.4 & 70.8 & $-36.4$ \\
PIQA            & 60.3 & 78.0 & $-17.8$ \\
WinoGrande      & 58.1 & 65.6 & $-7.5$ \\
\bottomrule
\end{tabular}
}
\caption{The average 0-shot accuracy (in percentage) is presented for both symbol-style and cloze-style formats, calculated across all evaluated models. The term $\Delta$ represents the difference in accuracy between the symbol and cloze formats, expressed in percentage points. For each task, a statistical test is conducted on the set of $\Delta$ values obtained from all evaluated models. In all tasks, the null hypothesis, $H_0: \Delta = 0$, is rejected with $p < 0.001$.}
\label{tab:3_0shot_sym_vs_clz}
\end{table}

\begin{figure}[!t]
\centering
\begin{footnotesize}

\begin{minipage}[t]{\linewidth}
\centering
\includegraphics[page=5,width=\linewidth]{figures/fewshot.pdf}
\end{minipage}

\vspace{0.4\baselineskip}

\begin{minipage}[t]{0.48\linewidth}
\centering
\includegraphics[page=1,width=\linewidth]{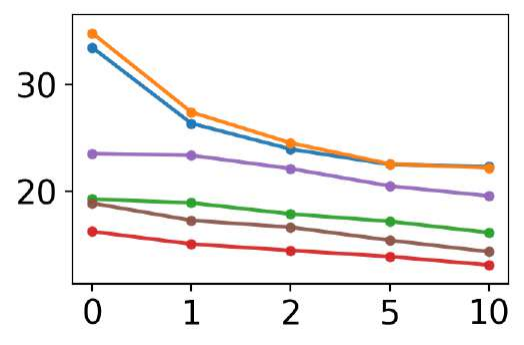}
(a) MMLU.
\end{minipage}
\hfill
\begin{minipage}[t]{0.48\linewidth}
\centering
\includegraphics[page=2,width=\linewidth]{figures/fewshot.pdf}
(b) ARC-Easy.
\end{minipage}

\vspace{0.5\baselineskip}
\begin{minipage}[t]{0.48\linewidth}
\centering
\includegraphics[page=3,width=\linewidth]{figures/fewshot.pdf}
(c) HellaSwag.
\end{minipage}
\hfill
\begin{minipage}[t]{0.48\linewidth}
\centering
\includegraphics[page=4,width=\linewidth]{figures/fewshot.pdf}
(d) PIQA.
\end{minipage}
\end{footnotesize}
\caption{Effect of in-context demonstrations on format mismatch. The vertical axis shows the performance difference between the symbol and cloze formats (symbol minus cloze, in percentage points), while the horizontal axis indicates the number of in-context demonstration examples.}
\label{fig:fewshot_delta}
\end{figure}

% tables/4_human_lable.tex
\begin{table*}[!t]
\centering

\footnotesize
\setlength{\tabcolsep}{4pt}
\renewcommand{\arraystretch}{1.1}
\resizebox{0.95\linewidth}{!}{%
\begin{tabularx}{\textwidth}{
    >{\centering\arraybackslash}m{0.07\textwidth}
    >{\centering\arraybackslash}m{0.1\textwidth}
    >{\RaggedRight\arraybackslash}m{0.33\textwidth}
    >{\RaggedRight\arraybackslash}X
}
\toprule
\textbf{Label} & \textbf{Typology} & \textbf{Description} & \textbf{Example} \\
\midrule\midrule

% Using * with multirow often handles vertical centering better in tabularx
\multirow{14.5}{*}{\textbf{Symbol}}
& \textbf{\makecell{Which}}
& An interrogative where the query is contextually dependent on the provided options for semantic grounding. 
& \renewcommand{\arraystretch}{1}
\begin{tabular}{l}
\textbf{Question:} Which item is used for protection from \\
\hspace{4em} chemical  splashing?\\
%\newline
  \textbf{Options:} A) compass \; B) hand lens \; C) microscope \; \\
\hspace{4em}   D) safety goggles 
\end{tabular}
  \\
\cmidrule{2-4}

& \textbf{\makecell{Short\\Answer}}
& A direct interrogative (non-\emph{which} type) where answer options consist of atomic entities, single words, or short phrases.
& \renewcommand{\arraystretch}{1}
\begin{tabular}{l}
  \textbf{Question:} What contains seeds?\\
  %\newline
  \textbf{Options:} A) human \; B) pumpkin \; C) soda can \; D) leaf 
  \end{tabular}
  \\
\cmidrule{2-4}

& \textbf{\makecell{Multi\\Sentence}}
& An interrogative where options are full declarative sentences that maintain independent syntactic structures from the question.
& \renewcommand{\arraystretch}{1}
\begin{tabular}{l}
\textbf{Question:} How does ice change the shape of rocks?\\
% \newline
  \textbf{Options:} A) It dissolves the rocks by pooling on surfaces. \; \\
  \hspace{4em} B) It breaks the rocks by expanding in openings. \; \\
  \hspace{4em} C) It smooths the rocks by colliding with them. \; \\
  \hspace{4em} D) It moves the rocks by pressing on them. 
    \end{tabular} \\
\cmidrule{2-4}

& \textbf{Imperative}
& A prompt formulated as a functional command or instruction, typically requiring problem-solving or calculation.
& \renewcommand{\arraystretch}{1}
\begin{tabular}{l}
\textbf{Question:} Solve the equation: $3x + 5 = 14$.\\
%\newline
  \textbf{Options:} A) $x=3$ \; B) $x=4$ \; C) $x=5$ \; D) $x=6$     
  \end{tabular} \\
\cmidrule{2-4}

& \textbf{\makecell{Anaphora\\Resolution}}
& A declarative statement containing an ambiguous anaphoric expression (e.g., ``this''), where options serve as candidate referents.
& \renewcommand{\arraystretch}{1}
 \begin{tabular}{l}
\textbf{Question:} This is most likely to be conserved:\\
%\newline
  \textbf{Options:} A) CO2 \; B) toilet paper \; C) a soda can \; \\
  \hspace{4em} D) styrofoam 
    \end{tabular} \\

\midrule

\multirow{8}{*}{\textbf{Cloze}}
& \textbf{\makecell{Blank\\Completion}}
& A prompt containing an explicit internal placeholder (\_\_) where the correct option completes a semantically coherent sentence.
& \renewcommand{\arraystretch}{1}
\begin{tabular}{l}
 \textbf{Question:} A/an \_\_\_\_\_\_ is reusable.\\
 %\newline
  \textbf{Options:} A) liquid soap \; B) dish towel \; C) band-aid \; \\
  \hspace{4em} D) apple 
    \end{tabular} \\
\cmidrule{2-4}

& \textbf{\makecell{Sentence\\Continuation}}
& A prompt consisting of an incomplete fragment or a full sentence that requires an option to form a logical narrative or sequential extension.
& \renewcommand{\arraystretch}{1}
\begin{tabular}{l}

\textbf{Question:} We see a bottle of face wash. We \\
%\newline
  \textbf{Options:} A) see a newspaper story. \; \\
    \hspace{4em} B) see four cookies in a pan. \; \\
      \hspace{4em} C) see a person holding face wash \\
      \hspace{5em}  then putting it on their face. \; \\
        \hspace{4em} D) see a tool on a table. 

  \end{tabular} \\
\bottomrule
\end{tabularx}
}
\caption{Human annotation guidelines with typologies and representative examples.}
\label{tab:human_label}
\end{table*}

\paragraph{Effect of in-context learning.}

The few-shot evaluation results shown in Figure~\ref{fig:fewshot_delta} indicate that the performance gains are significantly larger with the non-preferred evaluation format compared to the preferred one. Specifically, tasks that favor the symbol-based format show a greater impact of in-context learning during cloze-style evaluation than during symbol-based evaluation. Conversely, for tasks that favor the cloze-style format, performance improvements are much larger in that format, significantly reducing the issue of evaluation format mismatch. However, even with up to 10 demonstration examples, models often fail to match or remain noticeably below the baseline achieved with the optimal evaluation format. This suggests that while providing demonstration examples can partially alleviate evaluation format mismatch, models inherently struggle to interpret specific tasks effectively when faced with a non-preferred format. Motivated by these findings, we hypothesize that format mismatch is closely linked to the nature of the task. Thus, we investigate performance by selecting a task-specific evaluation format on a per-instance basis.

% \section{Methodology: Per-Instance Format Selection}
%\input{tables/4_human_label}

\section{Evaluation via Model-Preferred Formats}
\label{sec:model_preferred}
Based on the findings presented in Section~\ref{subsec:3_2_findings}, this section describes our approach to classifying each problem instance within a task and selecting the appropriate evaluation format. We introduce a lightweight encoder-based classifier that identifies the most suitable evaluation format for each problem instance.

\begin{figure*}[!t]
\centering
\includegraphics[page=1,width=\textwidth]{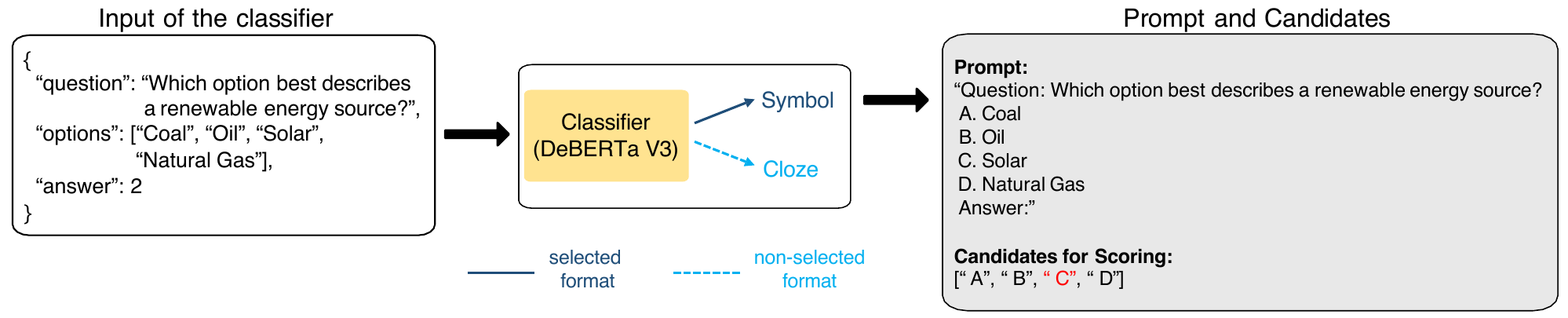}
\vspace*{-1\baselineskip}
\caption{Overview of instance-wise evaluation format selection. The classifier predicts the preferred format (\texttt{symbol} or \texttt{cloze}) for each instance. Based on the selected format, both the prompt fed into the LLM and the candidate set used for scoring are determined.}
\label{fig:classifier}
\end{figure*}

\subsection{The Classifier}
% Figure~\ref{fig:classifier} \hl{illustrates the instance-wise evaluation format selection pipeline and the resulting evaluation process.} We use DeBERTaV3~\cite{hedebertav3} as a lightweight classifier to find the evaluation format that is most suitable for each problem instance. The classifier receives as input the question text, the corresponding answer options, and the index of the candidate answer. Based on this input, it predicts whether a symbol-based or cloze-style evaluation format is more appropriate for the given instance. The predicted format is then used to determine how the language model is prompted and evaluated for that specific problem. 

% To train the classifier, we fine-tune it on the training splits of the benchmarks listed in Table~\ref{tab:benchmarks}. Each training example is annotated with a target evaluation format, following the labeling strategies outlined in Section~\ref{subsec:4_2_label}. Fine-tuning is conducted for three epochs using the AdamW optimizer~\cite{loshchilov2018decoupled}, with a learning rate of $2 \times 10^{-5}$. We employ a batch size of 32, with gradient accumulation, linear learning rate scheduling, and bfloat16 precision.

Figure~\ref{fig:classifier} illustrates the instance-wise format selection and the resulting evaluation process. The classifier takes as input the question text, the corresponding answer options, and the index of the answer. Based on this input, it predicts whether a symbol-based or cloze-style evaluation format is more appropriate for the given instance. The predicted format is then used to determine how the language model is prompted and evaluated for that specific problem. 

For the classifier, we fine-tune DeBERTaV3~\cite{hedebertav3} on the training splits of the benchmarks listed in Table~\ref{tab:benchmarks}. Each training example is annotated with a target evaluation format, following the labeling strategies outlined in Section~\ref{subsec:4_2_label}. Fine-tuning is conducted for three epochs using the AdamW\cite{loshchilov2018decoupled} optimizer, with a learning rate of $2 \times 10^{-5}$. We employ a batch size of 32, with gradient accumulation, a linear learning rate scheduling, and bfloat16 precision.

\subsection{Labeling Strategies for Training the Classifier}
\label{subsec:4_2_label}
The selection of labels (i.e., symbol or cloze) for fine-tuning instances is a crucial factor in testing our hypothesis: that each problem in a task has an evaluation format that is better suited to its nature. To systematically explore this, we employ two different labeling strategies for fine-tuning the classifier. The first strategy assesses whether performance improves when the evaluation format aligns with the human-judged one for each problem instance. The second examines whether preferences for specific formats are instead influenced by latent properties encoded in the target model that human annotators may not easily recognize.

\paragraph{Human-annotated labels.}
We initially hypothesized that an LLM's preference for a specific evaluation format is fundamentally rooted in the linguistic characteristics of the task. Given that these models are trained on vast human-generated corpora, we posit that their latent formatting preferences mirror human linguistic heuristics. Specifically, we expect models to favor formats that preserve the natural syntactic flow or logical structure inherent to the task type.

In our human-supervised training process, we manually annotate a subset of the training data to identify the most appropriate evaluation format for each problem instance. Specifically, for each benchmark, we randomly select 200 instances from the training set and assign a label indicating whether a symbol-based or cloze-style evaluation format is more suitable. To ensure that the labeling is deterministic and reproducible, human annotation follows strict syntactic criteria. We analyze the syntactic characteristics of each task to determine an evaluation format that aligns well with these properties. Table~\ref{tab:human_label} summarizes the detailed annotation guidelines and provides representative examples used by the authors, and all training data is labeled according to these guidelines.

The human labeling criteria ensure that cloze-style evaluation is used only when a single answer option can be seamlessly integrated into the question text to create a coherent statement. In contrast, symbol-based evaluation is applied when the correct resolution requires an explicit comparison among the answer options.

\begin{algorithm}[t]
\small
\caption{Labeling rule for problem instance $x$ with a margin threshold $\delta$.}
\label{alg:per_model_label}
\begin{algorithmic}[1]
\REQUIRE Correctness indicators $a^{\mathrm{sym}}, a^{\mathrm{clz}}\in\{0,1\}$ (1=correct, 0=incorrect), gold probabilities $p^{\mathrm{sym}}_{\mathrm{ans}}$ and $ p^{\mathrm{clz}}_{\mathrm{ans}}$, margins $d^{\mathrm{sym}}$ and $d^{\mathrm{clz}}$, a threshold $\delta$
\IF{$a^{\mathrm{sym}}=0$ \AND $a^{\mathrm{clz}}=0$}
    \STATE \textbf{abstain}
\ELSIF{$a^{\mathrm{sym}}=1$ \AND $a^{\mathrm{clz}}=0$}
    \IF{$p^{\mathrm{sym}}_{\mathrm{ans}} > p^{\mathrm{clz}}_{\mathrm{ans}}$}
        \STATE \textbf{return} \texttt{symbol}
    \ELSE
        \STATE \textbf{abstain}
    \ENDIF
\ELSIF{$a^{\mathrm{sym}}=0$ \AND $a^{\mathrm{clz}}=1$}
    \IF{$p^{\mathrm{clz}}_{\mathrm{ans}} > p^{\mathrm{sym}}_{\mathrm{ans}}$}
        \STATE \textbf{return} \texttt{cloze}
    \ELSE
        \STATE \textbf{abstain}
    \ENDIF
\ELSE[$a^{\mathrm{sym}}=1$ and $a^{\mathrm{clz}}=1$]
    \IF{$d^{\mathrm{sym}} - d^{\mathrm{clz}} \ge \delta$}
        \STATE \textbf{return} \texttt{symbol}
    \ELSIF{$d^{\mathrm{clz}} - d^{\mathrm{sym}} \ge \delta$}
        \STATE \textbf{return} \texttt{cloze}
    \ELSE
        \STATE \textbf{abstain}
    \ENDIF
\ENDIF
\end{algorithmic}
\label{algorithm:per_model}
\end{algorithm}

% \begin{algorithm}[H]
% \small
% \caption{Triad aggregation by strict majority voting.}
% \label{alg:triad_vote}
% \begin{algorithmic}[1]
% \REQUIRE A triad of label-source models $\mathcal{M}=\{m_1,m_2,m_3\}$, instance $x$, threshold $\delta$
% \FOR{$i\in\{1,2,3\}$}
%     \STATE Obtain $(a^{\mathrm{sym}}_i,a^{\mathrm{clz}}_i,\mathbf{p}^{\mathrm{sym}}_i,\mathbf{p}^{\mathrm{clz}}_i)$ for $x$ under model $m_i$
%     \STATE Compute $p^{\mathrm{sym}}_{i,\mathrm{ans}}, p^{\mathrm{clz}}_{i,\mathrm{ans}}$ and margins $d^{\mathrm{sym}}_i, d^{\mathrm{clz}}_i$
%     \STATE $\ell_i \leftarrow$ Algorithm~\ref{alg:per_model_label}$(a^{\mathrm{sym}}_i,a^{\mathrm{clz}}_i,p^{\mathrm{sym}}_{i,\mathrm{ans}},p^{\mathrm{clz}}_{i,\mathrm{ans}},d^{\mathrm{sym}}_i,d^{\mathrm{clz}}_i,\delta)$
%     \IF{$\ell_i$ is \textbf{abstain}}
%         \STATE \textbf{return} \textbf{abstain} \COMMENT{discard if any model abstains}
%     \ENDIF
% \ENDFOR
% \STATE \textbf{return} $\mathrm{MajorityVote}(\ell_1,\ell_2,\ell_3)$ \COMMENT{$\ell_i\in\{\texttt{symbol},\texttt{cloze}\}$}
% \end{algorithmic}
% \label{algorithm:triad}
% \end{algorithm}

\paragraph{Model-generated labels.}

To test the hypothesis that models may implicitly encode format preferences that are challenging for humans to identify explicitly, we create instance-level supervision signals by directly comparing the \texttt{symbol} and \texttt{cloze} formats with the target model on the training data. For a given problem instance $x$, we evaluate both formats with the model and obtain (i) a binary correctness indicator $a^{\mathrm{sym}}, a^{\mathrm{clz}}\in\{0,1\}$, where $1$ indicates correct and $0$ incorrect, and (ii) the full option probability distributions $\mathbf{p}^{\mathrm{sym}}$  and $\mathbf{p}^{\mathrm{clz}}$. The label of $x$ is determined by considering both the correctness and the confidence signals derived from the predicted probability distributions, as outlined in Algorithm~\ref{alg:per_model_label}.

In Algorithm~\ref{alg:per_model_label}, \textbf{abstain} denotes that we do not assign a label to the instance, and we exclude it from training. Instances where the model fails to accurately answer both formats are excluded from training. If the model correctly answers exactly one format, we consider the predicted probabilities of the correct answer for each format, denoted as $\mathbf{p}^{\mathrm{sym}}_{\mathrm{ans}}$ and $\mathbf{p}^{\mathrm{clz}}_{\mathrm{ans}}$, as confidence signals. The label is assigned based on a comparison of these probabilities. In cases where both formats are answered correctly, we compare their confidence margins, $d^{\mathrm{sym}}$ and $d^{\mathrm{clz}}$, which are defined as the difference between the highest and second-highest predicted probabilities among the options for each format. A label is assigned only if the absolute difference between the two margins exceeds a specified threshold, $\delta$.

% tables/5_rule_based.tex

\begin{table*}[!t]
\centering
\begin{minipage}{\linewidth}
\centering
\footnotesize
\setlength{\tabcolsep}{3pt}
\resizebox{\textwidth}{!}{%
\begin{tabular}{lccccccccc}
\toprule
\textbf{Model} &
\makecell{\textbf{ARC-}\\\textbf{Challenge}} &
\makecell{\textbf{ARC-}\\\textbf{Easy}} &
\makecell{\textbf{Commonsense-}\\\textbf{QA}} &
\textbf{MMLU} &
\makecell{\textbf{OpenBook-}\\\textbf{QA}} &
\textbf{HellaSwag} &
\makecell{\textbf{Wino-}\\\textbf{Grande}} &
\textbf{PIQA} &
\makecell{\textbf{SocialI-}\\\textbf{QA}}
\\
\midrule

% \multicolumn{10}{l}{\emph{Llama}~\cite{grattafiori2024llama3herdmodels}}\\
Llama 3.2 1B & 44.5 (+0.3) & 69.1 (+1.7) & 51.2 (-0.5) & 39.6 (-2.7) & 42.4 (-1.2) & 62.7 (-0.1) & 58.5 (-1.2) & 67.1 (-9.6) & 48.1 (+0.1) \\
Llama 3.1 8B & 76.7 (-2.7) & 90.5 (-1.3) & 70.4 (-0.3) & 57.4 (-7.3) & 58.2 (-17.4) & 78.4 (-0.1) & 70.3 (-0.2) & 70.8 (-10.4) & 62.7 (-0.1) \\

\midrule
% \multicolumn{10}{l}{\emph{Mistral}~\cite{jiang2023mistral7b}}\\
Mistral 7B v0.3 & 74.4 (-2.2) & 87.2 (-1.4) & 57.0 (-0.5) & 53.6 (-6.4) & 55.8 (-12.4) & 78.6 (+0.0) & 67.4 (-4.2) & 70.8 (-10.4) & 61.1 (-0.1) \\
Mistral Nemo Base 2407 & 77.0 (-4.1) & 90.1 (-0.9) & 55.6 (-2.0) & 57.0 (-6.9) & 57.8 (-15.4) & 80.0 (-0.0) & 72.1 (-1.5) & 70.2 (-12.0) & 64.9 (+0.1) \\

\midrule
% \multicolumn{10}{l}{\emph{Qwen 3}~\cite{yang2025qwen3}}\\
Qwen3 0.6B & 49.0 (-3.1) & 71.2 (-0.4) & 46.3 (-0.4) & 37.8 (-5.3) & 42.0 (-10.2) & 45.1 (+0.0) & 52.5 (-0.4) & 60.8 (-5.9) & 52.0 (-0.3) \\
Qwen3 8B Base & 87.3 (-5.2) & 94.8 (-2.4) & 85.7 (-0.2) & 63.5 (-12.5) & 61.0 (-24.0) & 75.6 (-0.2) & 68.4 (-0.1) & 75.3 (-4.1) & 78.8 (+0.2) \\

\bottomrule
\end{tabular}%
}
\caption{Model performance using a classifier trained on human-annotated labels. Values in parentheses indicate the performance gain relative to the baseline with the preferred format ($\max(\text{baseline}_{\texttt{symbol}}, \text{baseline}_{\texttt{cloze}})$).}

\label{tab:5_rule_based}
\end{minipage}

% tables/5_model_leveraged_full.tex

\newcommand{\twoscore}[2]{\makecell[c]{#1\\(#2)}}

\begin{minipage}{\linewidth}
\centering
\footnotesize
\setlength{\tabcolsep}{2pt}
\renewcommand{\arraystretch}{1.05}
\begin{adjustbox}{max width=\textwidth}
\begin{tabular}{ll*{9}{c}}
\toprule
\textbf{Model} & \textbf{Classifier} &
\makecell{\textbf{ARC-}\\\textbf{Challenge}} &
\makecell{\textbf{ARC-}\\\textbf{Easy}} &
\makecell{\textbf{Commonsense-}\\\textbf{QA}} &
\textbf{MMLU} &
\makecell{\textbf{OpenBook-}\\\textbf{QA}} &
\textbf{HellaSwag} &
\makecell{\textbf{Wino-}\\\textbf{Grande}} &
\textbf{PIQA} &
\makecell{\textbf{SocialI-}\\\textbf{QA}}\\
\midrule

% \multicolumn{11}{l}{\emph{Llama}}\\

\multirow{2}{*}{Llama 3.2 1B} & Self-labeled
& \twoscore{46.3}{+2.0} & \twoscore{69.9}{+2.5} & \twoscore{52.1}{+0.4}
& \twoscore{43.8}{+1.5} & \twoscore{47.4}{+3.8} & \twoscore{71.8}{+9.0} 
& \twoscore{76.4}{+16.7} & \twoscore{88.4}{+11.7} & \twoscore{48.5}{+0.5}\\
& Majority-labeled
& \twoscore{43.9}{-0.4} & \twoscore{67.5}{+0.1} & \twoscore{51.6}{-0.1}
& \twoscore{41.1}{-1.1} & \twoscore{44.0}{+0.4} & \twoscore{71.8}{+9.0} 
& \twoscore{69.8}{+10.1} & \twoscore{88.4}{+11.7} & \twoscore{48.1}{+0.1}\\
\cmidrule(lr){2-11}
\noalign{\vskip -1.5ex}
\addlinespace
\multirow{2}{*}{Llama 3.1 8B} & Self-labeled
& \twoscore{80.1}{+0.7} & \twoscore{92.1}{+0.3} & \twoscore{71.3}{+0.5}
& \twoscore{63.9}{-0.8} & \twoscore{76.6}{+1.0} & \twoscore{83.6}{+5.1} 
& \twoscore{78.5}{+8.0} & \twoscore{91.0}{+9.7} & \twoscore{62.7}{-0.0}\\
& Majority-labeled
& \twoscore{79.7}{+0.3} & \twoscore{92.0}{+0.1} & \twoscore{71.0}{+0.2}
& \twoscore{63.6}{-1.1} & \twoscore{76.6}{+1.0} & \twoscore{83.6}{+5.1} 
& \twoscore{76.2}{+5.6} & \twoscore{90.9}{+9.7} & \twoscore{62.6}{-0.1}\\

% \multicolumn{11}{l}{\emph{Mistral}}\\

\midrule
\multirow{2}{*}{Mistral 7B v0.3} & Self-labeled
& \twoscore{76.8}{+0.2} & \twoscore{88.6}{+0.0} & \twoscore{59.4}{+1.9}
& \twoscore{58.7}{-1.3} & \twoscore{68.8}{+0.6} & \twoscore{83.0}{+4.5} 
& \twoscore{78.5}{+6.9} & \twoscore{88.8}{+7.6} & \twoscore{60.8}{-0.3}\\
& Majority-labeled
& \twoscore{76.6}{-0.0} & \twoscore{88.4}{-0.2} & \twoscore{57.2}{-0.2}
& \twoscore{58.5}{-1.5} & \twoscore{68.4}{+0.2} & \twoscore{82.9}{+4.3} 
& \twoscore{72.7}{+1.1} & \twoscore{88.6}{+7.4} & \twoscore{61.1}{-0.1}\\
\cmidrule(lr){2-11}
\noalign{\vskip -1.5ex}
\addlinespace
\multirow{2}{*}{\makecell{Mistral Nemo Base 2407}} & Self-labeled
& \twoscore{81.0}{-0.1} & \twoscore{91.0}{-0.1} & \twoscore{58.6}{+1.1}
& \twoscore{62.6}{-1.3} & \twoscore{73.0}{-0.2} & \twoscore{84.3}{+4.3} 
& \twoscore{78.5}{+4.8} & \twoscore{89.5}{+7.3} & \twoscore{64.7}{-0.2}\\
& Majority-labeled
& \twoscore{80.7}{-0.3} & \twoscore{91.0}{-0.1} & \twoscore{55.9}{-1.6}
& \twoscore{62.3}{-1.6} & \twoscore{73.4}{+0.2} & \twoscore{84.2}{+4.3} 
& \twoscore{67.7}{-5.9} & \twoscore{89.2}{+7.0} & \twoscore{64.9}{+0.1}\\

% \multicolumn{11}{l}{\emph{Qwen}}\\

\midrule
\multirow{2}{*}{Qwen3 0.6B} & Self-labeled
& \twoscore{52.3}{+0.3} & \twoscore{71.9}{+0.3} & \twoscore{48.4}{+1.7}
& \twoscore{42.0}{-1.0} & \twoscore{53.0}{+0.8} & \twoscore{58.1}{+13.1} 
& \twoscore{81.2}{+28.3} & \twoscore{82.6}{+15.9} & \twoscore{52.0}{-0.3}\\
& Majority-labeled
& \twoscore{52.0}{+0.0} & \twoscore{71.6}{+0.0} & \twoscore{46.7}{+0.0}
& \twoscore{41.4}{-1.6} & \twoscore{52.2}{+0.0} & \twoscore{58.1}{+13.1} 
& \twoscore{72.7}{+19.8} & \twoscore{82.5}{+15.8} & \twoscore{52.0}{-0.3}\\
\cmidrule(lr){2-11}
\noalign{\vskip -1.5ex}
\addlinespace
\multirow{2}{*}{Qwen3 8B Base} & Self-labeled
& \twoscore{92.5}{+0.0} & \twoscore{97.3}{+0.1} & \twoscore{86.2}{+0.3}
& \twoscore{73.5}{-2.5} & \twoscore{84.2}{-0.8} & \twoscore{81.2}{+5.4} 
& \twoscore{71.3}{+2.8} & \twoscore{88.2}{+8.8} & \twoscore{78.8}{+0.2}\\
& Majority-labeled
& \twoscore{92.5}{+0.0} & \twoscore{97.3}{+0.1} & \twoscore{86.2}{+0.2}
& \twoscore{73.5}{-2.5} & \twoscore{84.4}{-0.6} & \twoscore{81.2}{+5.4} 
& \twoscore{68.1}{-0.4} & \twoscore{87.8}{+8.3} & \twoscore{78.8}{+0.2}\\

\bottomrule
\end{tabular}
\end{adjustbox}

\caption{Model performance using the classifier trained on model-generated labels. Values in parentheses indicate the performance gain relative to the baseline with the preferred format ($\max(\text{baseline}_{\texttt{symbol}}, \text{baseline}_{\texttt{cloze}})$).}
\label{tab:model_leveraged}
\end{minipage}

\begin{minipage}{\linewidth}

\end{minipage}

\end{table*}

%%%%%%%%%%%%%%%%%%%%%%%%%%%%%%%%%

\begin{table}{
\centering
\resizebox{\linewidth}{!}{%
\begin{tabular}{lcccc}
\toprule
& \multicolumn{2}{c}{\textbf{Human-annotated}} & \multicolumn{2}{c}{\textbf{Self-labeled}} \\
\cmidrule(lr){2-3} \cmidrule(lr){4-5}
\textbf{Task}
& \textbf{\makecell{\# Models with \\$\Delta \ge 0$}} 
& \textbf{\makecell{Mean \\$\Delta$}}
& \textbf{\makecell{\# Models with \\$\Delta \ge 0$}} 
& \textbf{\makecell{Mean \\$\Delta$}} \\
\midrule
ARC-Challenge   &  1/22 & $-3.53$ & 20/22 & $+0.25$ \\
ARC-Easy        &  1/22 & $-1.44$ & 21/22 & $+0.37$ \\
CommonsenseQA   &  4/22 & $-0.25$ & 21/22 & $+0.71$ \\
MMLU            &  0/22 & $-7.89$ &  3/22 & $-1.21$ \\
OpenBookQA      &  0/22 & $-16.96$ & 18/22 & $+0.67$ \\
HellaSwag       & 11/22 & $-0.28$ & 21/22 & $+5.66$ \\
WinoGrande      &  2/22 & $-1.18$ & 21/22 & $+6.97$ \\
PIQA            &  1/22 & $-6.19$ & 21/22 & $+7.40$ \\
SocialIQA       & 13/22 & $+0.05$ & 16/22 & $+0.07$ \\

\bottomrule
\end{tabular}%
}
\caption{
Task-level performance changes using the classifier trained with human-annotated labels or self-generated labels. 
Here, $\Delta$ denotes the performance gain relative to the baseline with the preferred format 
($\max(\text{baseline}_{\texttt{symbol}}, \text{baseline}_{\texttt{cloze}})$).
``\# Models with $\Delta \ge 0$'' counts models whose performance with the classifier matches or exceeds the baseline with the preferred format.
}
\label{tab:5_per_task}
}
\end{table}

\subsection{Training the Classifier}
\label{subsec:4_3_training}
We begin by training a classifier for a target model based on labels derived from the signals of that model. Specifically, the input to the classifier consists of the raw text instances from the training split of each evaluation task, paired with a label derived from either human annotation or model-generated signals. The specific methodologies for these labeling techniques are detailed in Section~\ref{subsec:4_2_label}. We then assess the target model's performance according to the predictions made by the classifier. 

Additionally, to investigate whether the signals are specific to individual models or are more generally shared among different decoder-based models, we select three decoder-based target models to generate training labels through a majority voting process. In this process, each of the three selected models independently assigns a label to each problem instance using the same labeling rule (as outlined in Algorithm~\ref{alg:per_model_label}). The final label is determined by the majority of non-abstaining votes; instances that do not have a clear majority are discarded.

\section{Evaluation Results}
\label{sec:eval}
In this section, we present the evaluation results for instance-wise format selection using the classifier fine-tuned according to the method introduced in Section~\ref{sec:model_preferred}. To determine whether the classifier demonstrates task-specific overfitting, we will also evaluate its performance on the SocialIQA dataset~\cite{sap2019social}.

\subsection{Human-Annotated Labels}
\label{subsec:human_eval}
Table~\ref{tab:5_rule_based} and Table~\ref{tab:5_per_task} present evaluation results for a classifier fine-tuned on human-annotated labels (For detailed evaluation results, please see Appendix~\ref{app:human_label}). In nearly all models and tasks, this approach leads to consistent performance degradation. These findings indicate that format mismatch is not primarily determined by intuitively salient problem properties, or that such properties are too nuanced to be reliably captured by rule-based labeling. In practice, tasks perceived as naturally suited to sentence completion are sometimes better handled using symbol-based formats. Conversely, questions that appear to require explicit comparison across options can often be more effectively solved using the cloze format.

These counterintuitive behaviors can be attributed to complex interactions between the model’s training objectives and its data exposure during pretraining and instruction tuning. Decoder-only LLMs are trained autoregressively to generate natural continuations, which biases them toward cloze-style sentence-completion formats (e.g., HellaSwag). This bias depends heavily on the formats in which similar tasks were predominantly encountered during training. In contrast, tasks that appear symbol-friendly from a human perspective may still favor cloze formats if the model has been extensively exposed to question–answer patterns that resemble cloze-style inputs.

Overall, these results suggest that format preferences cannot be reliably inferred from humans. Instead, directly extracting and using preference signals from model behavior offers a more effective approach to uncovering and leveraging the latent capabilities of decoder-based LLMs.

\subsection{Model-Generated Labels}
\label{subsec:model_eval}
Table~\ref{tab:model_leveraged} and Table~\ref{tab:5_per_task} present evaluation results based on selecting the evaluation format with a classifier fine-tuned on labels derived from each model (self-labeled). Labels are assigned following the procedure in Algorithm~\ref{alg:per_model_label}, with a margin threshold of $\delta = 0.2$. In the majority-voting setting (majority-labeled), labels are constructed from the predictions of three mid-sized models, each representing a distinct model family: Llama3.1 8B, Mistral 7B, and Qwen3 8B-Base. Full results across all evaluated models are reported in Appendix~\ref{app:model_leveraged}.

Across benchmarks that predominantly favor the cloze format, such as HellaSwag, PIQA, and Winogrande, the proposed approach produces substantial performance improvements, often surpassing the baseline with the preferred format. Conversely, on benchmarks favoring the symbol format, the proposed method maintains parity with the original preferred format and does not result in noticeable performance degradation.

These results demonstrate that even within tasks typically considered continuation-oriented, a significant subset of instances displays a clear preference for the symbol format. Accurately identifying these instances enables the model to better realize its latent capabilities. Although sentence-completion tasks generally align with the autoregressive training objective and tend to favor cloze-style inputs, format preference varies at the problem instance level based on characteristics and the distribution of related patterns encountered during training the target model. Therefore, evaluating all instances using a single, fixed format may underestimate the model’s actual performance potential.

To determine whether these preference signals generalize beyond individual models, another encoder is trained using labels derived from majority voting across three mid-sized models. Even in this aggregated context, consistent and significant gains are observed on cloze-preferring benchmarks. As shown in Table~\ref{tab:model_leveraged}, the performance difference between the cases of self-labeled and majority labeled is not significant.  These findings suggest that format preferences are not solely model-specific artifacts but instead reflect shared, task-dependent tendencies inherent to decoder-based LLMs. Leveraging aggregated signals enables robust, transferable format selection across different models.

\section{Conclusion}
This study investigates how evaluation formats, specifically symbol-based and cloze-style, systematically influence LLM performance in MCQA. The observed discrepancies are shown to be systematic, arising from task-level linguistic features and model inductive biases. Natural language continuation tasks tend to benefit from cloze-style formats, whereas tasks that require joint comparison of options are better suited to symbol-based evaluation. The analysis indicates that manual heuristics frequently fail to identify the optimal format. By using latent preference signals derived from model behavior, it is possible to achieve precise, instance-level format selection. This approach results in substantial performance improvements, particularly in completion-oriented benchmarks, and demonstrates generalizability across various model families. Fixed-format evaluation practices currently in use often underestimate LLM capabilities. For an accurate assessment of model performance, evaluation protocols should consider how models internally represent and respond to different interaction formats.

\section*{Impact Statement}
This paper presents work whose goal is to advance the field of large language model evaluation, with a particular focus on multiple-choice question answering (MCQA) evaluation methodologies. The work is primarily methodological in nature and aims to improve the reliability and interpretability of model evaluation. This line of research may further lead to a deeper understanding of the mechanisms through which intrinsic preference signals emerge within models, as well as to the development of training-free methods for effectively incorporating such preferences into evaluation. As such, it does not directly raise any societal or ethical concerns that warrant special consideration.

\section*{Acknowledgements}
This work was partially supported by the National Research Foundation of Korea (NRF) under Grant No. RS-2023-00222663 (Center for Optimizing Hyperscale AI Models and Platforms), and by the Institute for Information and Communications Technology Promotion (IITP) under Grant No. 2018-0-00581 (CUDA Programming Environment for FPGA Clusters) and No. RS-2025-02304554 (Efficient and Scalable Framework for AI Heterogeneous Cluster Systems), all funded by the Ministry of Science and ICT (MSIT) of Korea. It was also partially supported by the Korea Health Industry Development Institute (KHIDI) under Grant No. RS-2025-25454559 (Frailty Risk Assessment and Intervention Leveraging Multimodal Intelligence for Networked Deployment in Community Care), funded by the Ministry of Health and Welfare (MOHW) of Korea. Additional support was provided by the BK21 Plus Program for Innovative Data Science Talent Education (Department of Data Science, Seoul National University, No. 5199990914569) and the BK21 FOUR Program for Intelligent Computing (Department of Computer Science and Engineering, Seoul National University, No. 4199990214639), both funded by the Ministry of Education (MOE) of Korea. This work was also partially supported by the Artificial Intelligence Industrial Convergence Cluster Development Project, funded by the MSIT and Gwangju Metropolitan City. Research facilities were provided by the Institute of Computer Technology (ICT) at Seoul National University.

\newpage
\bibliography{custom}

@article{achiam2023gpt,
  title={Gpt-4 technical report},
  author={Achiam, Josh and Adler, Steven and Agarwal, Sandhini and Ahmad, Lama and Akkaya, Ilge and Aleman, Florencia Leoni and Almeida, Diogo and Altenschmidt, Janko and Altman, Sam and Anadkat, Shyamal and others},
  journal={arXiv preprint arXiv:2303.08774},
  year={2023}
}

@article{team2023gemini,
  title={Gemini: a family of highly capable multimodal models},
  author={Team, Gemini and Anil, Rohan and Borgeaud, Sebastian and Alayrac, Jean-Baptiste and Yu, Jiahui and Soricut, Radu and Schalkwyk, Johan and Dai, Andrew M and Hauth, Anja and Millican, Katie and others},
  journal={arXiv preprint arXiv:2312.11805},
  year={2023}
}

@misc{jiang2023mistral7b,
      title={Mistral 7B}, 
      author={Albert Q. Jiang and Alexandre Sablayrolles and Arthur Mensch and Chris Bamford and Devendra Singh Chaplot and Diego de las Casas and Florian Bressand and Gianna Lengyel and Guillaume Lample and Lucile Saulnier and Lélio Renard Lavaud and Marie-Anne Lachaux and Pierre Stock and Teven Le Scao and Thibaut Lavril and Thomas Wang and Timothée Lacroix and William El Sayed},
      year={2023},
      eprint={2310.06825},
      archivePrefix={arXiv},
      primaryClass={cs.CL},
      url={https://arxiv.org/abs/2310.06825}, 
}

@article{grattafiori2024llama3herdmodels,
  title={The llama 3 herd of models},
  author={Grattafiori, Aaron and Dubey, Abhimanyu and Jauhri, Abhinav and Pandey, Abhinav and Kadian, Abhishek and Al-Dahle, Ahmad and Letman, Aiesha and Mathur, Akhil and Schelten, Alan and Vaughan, Alex and others},
  journal={arXiv preprint arXiv:2407.21783},
  year={2024}
}

@inproceedings{alzahrani2024benchmarks,
  title={When benchmarks are targets: Revealing the sensitivity of large language model leaderboards},
  author={Alzahrani, Norah and Alyahya, Hisham and Alnumay, Yazeed and Alrashed, Sultan and Alsubaie, Shaykhah and Almushayqih, Yousef and Mirza, Faisal and Alotaibi, Nouf and Al-Twairesh, Nora and Alowisheq, Areeb and others},
  booktitle={Proceedings of the 62nd Annual Meeting of the Association for Computational Linguistics (Volume 1: Long Papers)},
  pages={13787--13805},
  year={2024}
}

@article{brown2020language,
  title={Language models are few-shot learners},
  author={Brown, Tom and Mann, Benjamin and Ryder, Nick and Subbiah, Melanie and Kaplan, Jared D and Dhariwal, Prafulla and Neelakantan, Arvind and Shyam, Pranav and Sastry, Girish and Askell, Amanda and others},
  journal={Advances in neural information processing systems},
  volume={33},
  pages={1877--1901},
  year={2020}
}

@inproceedings{lyu2024beyond,
  title={Beyond probabilities: Unveiling the misalignment in evaluating large language models},
  author={Lyu, Chenyang and Wu, Minghao and Aji, Alham},
  booktitle={Proceedings of the 1st Workshop on Towards Knowledgeable Language Models (KnowLLM 2024)},
  pages={109--131},
  year={2024}
}

@article{touvron2023llama,
  title={Llama: Open and efficient foundation language models},
  author={Touvron, Hugo and Lavril, Thibaut and Izacard, Gautier and Martinet, Xavier and Lachaux, Marie-Anne and Lacroix, Timoth{\'e}e and Rozi{\`e}re, Baptiste and Goyal, Naman and Hambro, Eric and Azhar, Faisal and others},
  journal={arXiv preprint arXiv:2302.13971},
  year={2023}
}

@article{sutawika2025eleutherai,
  title={EleutherAI/lm-evaluation-harness: v0. 4.9},
  author={Sutawika, Lintang and Schoelkopf, Hailey and Gao, Leo and Abbasi, Baber and Biderman, Stella and Tow, Jonathan and Lovering, Charles and Phang, Jason and Thite, Anish and Wang, Thomas and others},
  journal={Zenodo}, year={2025}
}

@article{zheng2023large,
  title={Large language models are not robust multiple choice selectors},
  author={Zheng, Chujie and Zhou, Hao and Meng, Fandong and Zhou, Jie and Huang, Minlie},
  journal={arXiv preprint arXiv:2309.03882},
  year={2023}
}

@inproceedings{pezeshkpour2024large,
  title={Large language models sensitivity to the order of options in multiple-choice questions},
  author={Pezeshkpour, Pouya and Hruschka, Estevam},
  booktitle={Findings of the Association for Computational Linguistics: NAACL 2024},
  pages={2006--2017},
  year={2024}
}

@inproceedings{lu2022fantastically,
  title={Fantastically ordered prompts and where to find them: Overcoming few-shot prompt order sensitivity},
  author={Lu, Yao and Bartolo, Max and Moore, Alastair and Riedel, Sebastian and Stenetorp, Pontus},
  booktitle={Proceedings of the 60th Annual Meeting of the Association for Computational Linguistics (Volume 1: Long Papers)},
  pages={8086--8098},
  year={2022}
}

@inproceedings{sclarquantifying,
  title={Quantifying Language Models' Sensitivity to Spurious Features in Prompt Design or: How I learned to start worrying about prompt formatting},
  author={Sclar, Melanie and Choi, Yejin and Tsvetkov, Yulia and Suhr, Alane},
  booktitle={The Twelfth International Conference on Learning Representations}
}

@inproceedings{zong2024fool,
  title={Fool Your (Vision and) Language Model with Embarrassingly Simple Permutations},
  author={Zong, Yongshuo and Yu, Tingyang and Chavhan, Ruchika and Zhao, Bingchen and Hospedales, Timothy},
  booktitle={International Conference on Machine Learning},
  pages={62892--62913},
  year={2024},
  organization={PMLR}
}

@inproceedings{wei2024unveiling,
  title={Unveiling Selection Biases: Exploring Order and Token Sensitivity in Large Language Models},
  author={Wei, Sheng-Lun and Wu, Cheng-Kuang and Huang, Hen-Hsen and Chen, Hsin-Hsi},
  booktitle={Findings of the Association for Computational Linguistics ACL 2024},
  pages={5598--5621},
  year={2024}
}

@article{zhou2024large,
  title={Large language model (llm) for telecommunications: A comprehensive survey on principles, key techniques, and opportunities},
  author={Zhou, Hao and Hu, Chengming and Yuan, Ye and Cui, Yufei and Jin, Yili and Chen, Can and Wu, Haolun and Yuan, Dun and Jiang, Li and Wu, Di and others},
  journal={IEEE Communications Surveys \& Tutorials},
  volume={27},
  number={3},
  pages={1955--2005},
  year={2024},
  publisher={IEEE}
}

@inproceedings{liusie2024teacher,
  title={Teacher-student training for debiasing: General permutation debiasing for large language models},
  author={Liusie, Adian and Fathullah, Yassir and Gales, Mark},
  booktitle={Findings of the Association for Computational Linguistics: ACL 2024},
  pages={1376--1387},
  year={2024}
}

@inproceedings{choi2025mitigating,
  title={Mitigating selection bias with node pruning and auxiliary options},
  author={Choi, Hyeong Kyu and Xu, Weijie and Xue, Chi and Eckman, Stephanie and Reddy, Chandan K},
  booktitle={Proceedings of the 63rd Annual Meeting of the Association for Computational Linguistics (Volume 1: Long Papers)},
  pages={5190--5215},
  year={2025}
}

@inproceedings{
robinson2023leveraging,
title={Leveraging Large Language Models for Multiple Choice Question Answering},
author={Joshua Robinson and David Wingate},
booktitle={The Eleventh International Conference on Learning Representations },
year={2023},
url={https://openreview.net/forum?id=yKbprarjc5B}
}

@inproceedings{gu2025olmes,
  title={Olmes: A standard for language model evaluations},
  author={Gu, Yuling and Tafjord, Oyvind and Kuehl, Bailey and Haddad, Dany and Dodge, Jesse and Hajishirzi, Hannaneh},
  booktitle={Findings of the Association for Computational Linguistics: NAACL 2025},
  pages={5005--5033},
  year={2025}
}

@article{yang2025qwen3,
  title={Qwen3 technical report},
  author={Yang, An and Li, Anfeng and Yang, Baosong and Zhang, Beichen and Hui, Binyuan and Zheng, Bo and Yu, Bowen and Gao, Chang and Huang, Chengen and Lv, Chenxu and others},
  journal={arXiv preprint arXiv:2505.09388},
  year={2025}
}

@inproceedings{belinkov2019don,
  title={Don’t Take the Premise for Granted: Mitigating Artifacts in Natural Language Inference},
  author={Belinkov, Yonatan and Poliak, Adam and Shieber, Stuart M and Van Durme, Benjamin and Rush, Alexander M},
  booktitle={Proceedings of the 57th Annual Meeting of the Association for Computational Linguistics},
  pages={877--891},
  year={2019}
}

@inproceedings{feng2019misleading,
  title={Misleading Failures of Partial-input Baselines},
  author={Feng, Shi and Wallace, Eric and Boyd-Graber, Jordan},
  booktitle={Proceedings of the 57th Annual Meeting of the Association for Computational Linguistics},
  pages={5533--5538},
  year={2019}
}

@inproceedings{srikanth2022partial,
  title={Partial-input baselines show that NLI models can ignore context, but they don’t.},
  author={Srikanth, Neha and Rudinger, Rachel},
  booktitle={Proceedings of the 2022 Conference of the North American Chapter of the Association for Computational Linguistics: Human Language Technologies},
  pages={4753--4763},
  year={2022}
}

@article{cho2026choicesspeaklouderquestions,
  title={Choices Speak Louder than Questions},
  author={Cho, Gyeongje and So, Yeonkyoung and Lee, Jaejin},
  journal={arXiv preprint arXiv:2502.18798},
  year={2026},
  note = {To appear in ICLR 2026}
}

@article{clark2018think,
  title={Think you have solved question answering? try arc, the ai2 reasoning challenge},
  author={Clark, Peter and Cowhey, Isaac and Etzioni, Oren and Khot, Tushar and Sabharwal, Ashish and Schoenick, Carissa and Tafjord, Oyvind},
  journal={arXiv preprint arXiv:1803.05457},
  year={2018}
}

@inproceedings{zellers2019hellaswag,
  title={HellaSwag: Can a Machine Really Finish Your Sentence?},
  author={Zellers, Rowan and Holtzman, Ari and Bisk, Yonatan and Farhadi, Ali and Choi, Yejin},
  booktitle={Proceedings of the 57th Annual Meeting of the Association for Computational Linguistics},
  pages={4791--4800},
  year={2019}
}

@article{sakaguchi2021winogrande,
  title={Winogrande: An adversarial winograd schema challenge at scale},
  author={Sakaguchi, Keisuke and Bras, Ronan Le and Bhagavatula, Chandra and Choi, Yejin},
  journal={Communications of the ACM},
  volume={64},
  number={9},
  pages={99--106},
  year={2021},
  publisher={ACM New York, NY, USA}
}

@inproceedings{
hendrycks2021measuring,
title={Measuring Massive Multitask Language Understanding},
author={Dan Hendrycks and Collin Burns and Steven Basart and Andy Zou and Mantas Mazeika and Dawn Song and Jacob Steinhardt},
booktitle={International Conference on Learning Representations},
year={2021},
url={https://openreview.net/forum?id=d7KBjmI3GmQ}
}

@inproceedings{mihaylov2018can,
  title={Can a Suit of Armor Conduct Electricity? A New Dataset for Open Book Question Answering},
  author={Mihaylov, Todor and Clark, Peter and Khot, Tushar and Sabharwal, Ashish},
  booktitle={Proceedings of the 2018 Conference on Empirical Methods in Natural Language Processing},
  pages={2381--2391},
  year={2018}
}

@inproceedings{talmor2019commonsenseqa,
  title={Commonsenseqa: A question answering challenge targeting commonsense knowledge},
  author={Talmor, Alon and Herzig, Jonathan and Lourie, Nicholas and Berant, Jonathan},
  booktitle={Proceedings of the 2019 Conference of the North American Chapter of the Association for Computational Linguistics: Human Language Technologies, Volume 1 (Long and Short Papers)},
  pages={4149--4158},
  year={2019}
}

@inproceedings{bisk2020piqa,
  title={Piqa: Reasoning about physical commonsense in natural language},
  author={Bisk, Yonatan and Zellers, Rowan and Gao, Jianfeng and Choi, Yejin and others},
  booktitle={Proceedings of the AAAI conference on artificial intelligence},
  volume={34},
  number={05},
  pages={7432--7439},
  year={2020}
}

@inproceedings{sap2019social,
  title={Social IQa: Commonsense Reasoning about Social Interactions},
  author={Sap, Maarten and Rashkin, Hannah and Chen, Derek and Le Bras, Ronan and Choi, Yejin},
  booktitle={Proceedings of the 2019 Conference on Empirical Methods in Natural Language Processing and the 9th International Joint Conference on Natural Language Processing (EMNLP-IJCNLP)},
  pages={4463--4473},
  year={2019}
}

@inproceedings{hedebertav3,
  title={DeBERTaV3: Improving DeBERTa using ELECTRA-Style Pre-Training with Gradient-Disentangled Embedding Sharing},
  author={He, Pengcheng and Gao, Jianfeng and Chen, Weizhu},
  booktitle={The Eleventh International Conference on Learning Representations},
  year={2023}
}

@inproceedings{
loshchilov2018decoupled,
title={Decoupled Weight Decay Regularization},
author={Ilya Loshchilov and Frank Hutter},
booktitle={International Conference on Learning Representations},
year={2019},
url={https://openreview.net/forum?id=Bkg6RiCqY7},
}
\bibliographystyle{icml2026}
\newpage
\appendix
\onecolumn
\section{Details of the Models}
\label{app:models}
Table~\ref{tab:models} shows the all models evaluated in this paper. For all evaluation, we use a 16-bit (bfloat16) quantized model.

\begin{table}[htbp]
\centering
\resizebox{0.8\textwidth}{!}{%
\begin{tabular}{@{}lll@{}}
\toprule
\textbf{Qwen 3}\cite{yang2025qwen3} & \textbf{Llama 3.1 \& 3.2}\cite{grattafiori2024llama3herdmodels} & \textbf{Mistral}\cite{jiang2023mistral7b} \\ 
\midrule
Qwen3-0.6B-Base  & Llama-3.1-8B           & Mistral-7B-v0.3 \\
Qwen3-0.6B       & Llama-3.1-8B-Instruct  & Mistral-7B-Instruct-v0.3 \\
Qwen3-1.7B-Base  & Llama-3.2-1B           & Mistral-Nemo-Base-2407 (12B)\\
Qwen3-1.7B       & Llama-3.2-1B-Instruct  & Mistral-Nemo-Instruct-2407 (12B)\\
Qwen3-4B-Base    & Llama-3.2-3B           & Mistral-Small-24B-Base-2501 \\
Qwen3-4B         & Llama-3.2-3B-Instruct  & Mistral-Small-24B-Instruct-2501 \\
Qwen3-8B-Base    &                        &  \\
Qwen3-8B         &                        &  \\
Qwen3-14B-Base   &                        &  \\
Qwen3-14B        &                        &  \\
\bottomrule
\end{tabular}%
}

\caption{Models evaluated in our experiments.}
\label{tab:models}
\end{table}

% appendix_prompts.tex
% (Assumes in the main preamble you have defined:
%  - \begin{benchbox} ... \end{benchbox}  (tcolorbox)
%  - \begin{promptblock} ... \end{promptblock} (Verbatim monospace)

\section{Detailed Prompt Examples for Each Benchmark}
\label{app:prompts}
This section presents the complete set of prompt templates used for evaluation across all benchmarks. For the symbol format, we define task-specific prefixes and suffixes based on the characteristics of each task.

% ---------------- ARC ----------------
\noindent\textbf{ARC-Easy/Challenge}\cite{clark2018think}\par
\begin{benchbox}
\textbf{Symbol}\par
\begin{promptblock}
Prompt:
"Question: Which factor will most likely cause a person to develop a fever?
A. a leg muscle relaxing after exercise
B. a bacterial population in the bloodstream
C. several viral particles on the skin
D. carbohydrates being digested in the stomach
Answer:"

Candidates for Scoring:
[" A", " B", " C", " D"]
\end{promptblock}

\textbf{Cloze}\par
\begin{promptblock}
Prompt:
"Which factor will most likely cause a person to develop a fever?"

Candidates for Scoring:
[" a leg muscle relaxing after exercise",
 " a bacterial population in the bloodstream",
 " several viral particles on the skin",
 " carbohydrates being digested in the stomach"]
\end{promptblock}
\end{benchbox}

\vspace{0.9em}

\clearpage
% ---------------- CSQA ----------------
\noindent\textbf{CommonsenseQA}\cite{talmor2019commonsenseqa}\par
\begin{benchbox}
\textbf{Symbol}\par
\begin{promptblock}
Prompt:
"Question: Unlike young people, older people can do what?
A. talk to each other
B. become hysterical
C. chat with each other
D. grow shorter
E. take trips
Answer:"

Candidates for Scoring:
[" A", " B", " C", " D", " E"]
\end{promptblock}

\textbf{Cloze}\par
\begin{promptblock}
Prompt:
"Unlike young people, older people can do what?"

Candidates for Scoring:
[" talk to each other",
 " become hysterical",
 " chat with each other",
 " grow shorter",
 " take trips"]
\end{promptblock}
\end{benchbox}

\vspace{0.9em}

% ---------------- MMLU ----------------
\noindent\textbf{MMLU}\cite{hendrycks2021measuring}\par
\begin{benchbox}
\textbf{Symbol}\par
\begin{promptblock}
Prompt:
"Question: A high school science teacher fills a 1 liter bottle with pure nitrogen and seals the lid. The pressure is 1.70 atm, and the room temperature is 25°C. Which two variables will both increase the pressure of the system, if all other variables are held constant?
A. Increasing temperature, increasing moles of gas
B. Increasing temperature, increasing volume
C. Decreasing volume, decreasing temperature
D. Decreasing moles of gas, increasing volume
Answer:"

Candidates for Scoring:
[" A", " B", " C", " D"]
\end{promptblock}

\textbf{Cloze}\par
\begin{promptblock}
Prompt:
"A high school science teacher fills a 1 liter bottle with pure nitrogen and seals the lid. The pressure is 1.70 atm, and the room temperature is 25°C. Which two variables will both increase the pressure of the system, if all other variables are held constant?"

Candidates for Scoring:
[" Increasing temperature, increasing moles of gas",
 " Increasing temperature, increasing volume",
 " Decreasing volume, decreasing temperature",
 " Decreasing moles of gas, increasing volume"]
\end{promptblock}
\end{benchbox}

\vspace{0.9em}

\clearpage
% ---------------- OpenBookQA ----------------
\noindent\textbf{OpenBookQA}\cite{mihaylov2018can}\par
\begin{benchbox}
\textbf{Symbol}\par
\begin{promptblock}
Prompt:
"Question: Poison causes harm to which of the following?
A. a Tree
B. a robot
C. a house
D. a car
Answer:"

Candidates for Scoring:
[" A", " B", " C", " D"]
\end{promptblock}

\textbf{Cloze}\par
\begin{promptblock}
Prompt:
"Poison causes harm to which of the following?"

Candidates for Scoring:
[" a Tree",
 " a robot",
 " a house",
 " a car"]
\end{promptblock}
\end{benchbox}

\vspace{0.9em}

% ---------------- HellaSwag ----------------
\noindent\textbf{HellaSwag}\cite{zellers2019hellaswag}\par
\begin{benchbox}
\textbf{Symbol}\par
\begin{promptblock}
Prompt:
"Context: A helicopter flies in some people who then start playing paintball. they
A. run around obstacles and have a great time.
B. lose their shirt during the fight and run back the others.
C. shoot at each other while the helicopter drone who am looking on.
D. chase a bird in the sky while others walk around.
Completion:"

Candidates for Scoring:
[" A", " B", " C", " D"]
\end{promptblock}

\textbf{Cloze}\par
\begin{promptblock}
Prompt:
"A helicopter flies in some people who then start playing paintball. they"

Candidates for Scoring:
[" run around obstacles and have a great time.",
 " lose their shirt during the fight and run back the others.",
 " shoot at each other while the helicopter drone who am looking on.",
 " chase a bird in the sky while others walk around."]
\end{promptblock}
\end{benchbox}

\vspace{0.9em}

\clearpage
% ---------------- WinoGrande ----------------
\noindent\textbf{WinoGrande}\cite{sakaguchi2021winogrande}\par
\begin{benchbox}
\textbf{Symbol}\par
\begin{promptblock}
Prompt:
"Sentence: The spray cleaned the windows better than it cleaned the walls because the _ were non-porous.
A. windows
B. walls
Answer:"

Candidates for Scoring:
[" A", " B"]
\end{promptblock}

\textbf{Cloze}\par
\begin{promptblock}
Prompt:
"The spray cleaned the windows better than it cleaned the walls because the"

Candidates for Scoring:
[" windows were non-porous.",
 " walls were non-porous."]
\end{promptblock}
\end{benchbox}

\vspace{0.9em}

% ---------------- PIQA ----------------
\noindent\textbf{PIQA}\cite{bisk2020piqa}\par
\begin{benchbox}
\textbf{Symbol}\par
\begin{promptblock}
Prompt:
"Context: To clean rust stains off of the bath tub.
A. Use half a grapefruit and pepper. Sprinkle about a 1/4 cup of kosher pepper on a halved grapefruit.
B. Use half a grapefruit and salt. Sprinkle about a 1/4 cup of kosher salt on a halved grapefruit.
Completion:"

Candidates for Scoring:
[" A", " B"]
\end{promptblock}

\textbf{Cloze}\par
\begin{promptblock}
Prompt:
"To clean rust stains off of the bath tub."

Candidates for Scoring:
[" Use half a grapefruit and pepper. Sprinkle about a 1/4 cup of kosher pepper on a halved grapefruit.",
 " Use half a grapefruit and salt. Sprinkle about a 1/4 cup of kosher salt on a halved grapefruit."]
\end{promptblock}
\end{benchbox}

\vspace{0.9em}

% ---------------- SocialIQA ----------------
\noindent\textbf{SocialIQA}\cite{sap2019social}\par
\begin{benchbox}
\textbf{Symbol}\par
\begin{promptblock}
Prompt:
"Context: Sydney walked past a homeless woman asking for change but did not have any money they could give to her. Sydney felt bad afterwards.
Question: How would you describe Sydney?
A. sympathetic
B. like a person who was unable to help
C. incredulous
Answer:"

Candidates for Scoring:
[" A", " B", " C"]
\end{promptblock}

\textbf{Cloze}\par
\begin{promptblock}
Prompt:
"Sydney walked past a homeless woman asking for change but did not have any money they could give to her. Sydney felt bad afterwards.
How would you describe Sydney?"

Candidates for Scoring:
[" sympathetic",
 " like a person who was unable to help",
 " incredulous"]
\end{promptblock}
\end{benchbox}

\clearpage

\section{Detailed Baseline Results}
\label{app:baseline}
In this section, we provide the comprehensive baseline results referenced in Section~\ref{subsec:3_2_findings}. Table~\ref{tab:symbol_cloze_llama_3_2_1b}-~\ref{tab:symbol_cloze_qwen3_14b} present the performance of each model across all benchmarks. We report model performance on each task under both the \texttt{symbol} and \texttt{cloze} settings. For the symbol-based format, we measure performance using standard accuracy. For the cloze-style evaluation, we report length-normalized accuracy, computed by dividing the raw log-likelihood of each option by the number of tokens~\cite{brown2020language}.
For few-shot evaluation, results are averaged over five random seeds used to sample demonstration examples (308, 713, 777, 1,234, and 4,649)

% tables/app_baseline.tex

% ======================
% Llama
% ======================
\begin{table}[!htbp]
\centering
\small
\setlength{\tabcolsep}{4pt}
\begin{tabular}{p{0.20\textwidth} p{0.11\textwidth} rrrrr}
\toprule
\textbf{Task} & \textbf{Method} & \textbf{0-shot} & \textbf{1-shot} & \textbf{2-shot} & \textbf{5-shot} & \textbf{10-shot} \\
\midrule
ARC-Challenge & Symbol & 44.28 & 34.95 & 35.58 & 34.44 & 32.46 \\
 & Cloze & 35.49 & 34.90 & 35.80 & 37.30 & 38.16 \\
\midrule
ARC-Easy & Symbol & 67.38 & 49.90 & 47.12 & 46.03 & 37.93 \\
 & Cloze & 54.97 & 61.06 & 65.85 & 68.80 & 69.98 \\
 \midrule
CommonsenseQA & Symbol & 51.68 & 30.35 & 32.55 & 29.58 & 28.70 \\
 & Cloze & 39.23 & 47.70 & 55.48 & 58.07 & 61.38 \\
 \midrule
MMLU & Symbol & 42.25 & 34.16 & 34.02 & 33.37 & 32.27 \\
 & Cloze & 34.10 & 34.64 & 35.79 & 36.94 & 37.95 \\
 \midrule
OpenBookQA & Symbol & 43.60 & 33.16 & 37.00 & 34.76 & 34.04 \\
 & Cloze & 39.00 & 37.56 & 38.28 & 40.08 & 41.08 \\
 \midrule
HellaSwag & Symbol & 25.14 & 25.48 & 25.24 & 25.46 & 25.70 \\
 & Cloze & 62.72 & 62.41 & 62.67 & 63.37 & 63.43 \\
 \midrule
WinoGrande & Symbol & 49.49 & 49.87 & 50.21 & 49.88 & 49.91 \\
 & Cloze & 58.96 & 59.13 & 59.10 & 58.55 & 58.64 \\
 \midrule
PIQA & Symbol & 49.08 & 50.86 & 51.77 & 50.71 & 50.82 \\
 & Cloze & 76.66 & 75.39 & 75.56 & 76.30 & 76.31 \\
\bottomrule
\end{tabular}
\caption{Performance (\%) comparing Symbol and Cloze formats across $k$-shot settings ($k \in \{0,1,2,5,10\}$) for Llama-3.2-1B.}
\label{tab:symbol_cloze_llama_3_2_1b}
\end{table}

\begin{table}[!htbp]
\centering
\small
\setlength{\tabcolsep}{4pt}
\begin{tabular}{p{0.20\textwidth} p{0.11\textwidth} rrrrr}
\toprule
\textbf{Task} & \textbf{Method} & \textbf{0-shot} & \textbf{1-shot} & \textbf{2-shot} & \textbf{5-shot} & \textbf{10-shot} \\
\midrule
ARC-Challenge & Symbol & 54.61 & 50.65 & 53.11 & 53.86 & 54.47 \\
 & Cloze & 35.92 & 35.54 & 36.18 & 36.79 & 37.71 \\
\midrule
ARC-Easy & Symbol & 75.29 & 69.76 & 72.53 & 74.06 & 74.28 \\
 & Cloze & 61.95 & 60.54 & 64.11 & 66.61 & 68.06 \\
\midrule
CommonsenseQA & Symbol & 59.13 & 54.20 & 56.30 & 59.30 & 59.30 \\
 & Cloze & 43.41 & 50.06 & 54.94 & 56.61 & 59.07 \\
\midrule
MMLU & Symbol & 45.79 & 45.09 & 46.69 & 47.28 & 47.42 \\
 & Cloze & 35.07 & 34.96 & 35.73 & 36.76 & 37.60 \\
\midrule
OpenBookQA & Symbol & 59.80 & 54.88 & 56.84 & 58.28 & 58.00 \\
 & Cloze & 38.00 & 38.72 & 39.00 & 39.20 & 39.64 \\
\midrule
HellaSwag & Symbol & 25.19 & 31.44 & 35.53 & 35.72 & 35.15 \\
 & Cloze & 59.82 & 59.32 & 59.29 & 59.67 & 59.39 \\
\midrule
WinoGrande & Symbol & 50.28 & 51.68 & 52.49 & 52.75 & 52.23 \\
 & Cloze & 57.77 & 55.85 & 56.31 & 56.35 & 56.61 \\
\midrule
PIQA & Symbol & 51.74 & 60.50 & 62.55 & 62.37 & 59.91 \\
 & Cloze & 74.21 & 74.02 & 74.07 & 74.76 & 74.55 \\
\bottomrule
\end{tabular}
\caption{Performance (\%) comparing Symbol and Cloze formats across $k$-shot settings ($k \in \{0,1,2,5,10\}$) for Llama-3.2-1B-Instruct.}
\label{tab:symbol_cloze_llama_3_2_1b_inst}
\end{table}

\begin{table}[!htbp]
\centering
\small
\setlength{\tabcolsep}{4pt}
\begin{tabular}{p{0.20\textwidth} p{0.11\textwidth} rrrrr}
\toprule
\textbf{Task} & \textbf{Method} & \textbf{0-shot} & \textbf{1-shot} & \textbf{2-shot} & \textbf{5-shot} & \textbf{10-shot} \\
\midrule
ARC-Challenge & Symbol & 67.75 & 68.89 & 69.11 & 69.04 & 68.87 \\
 & Cloze & 38.40 & 39.05 & 39.80 & 41.76 & 43.29 \\
\midrule
ARC-Easy & Symbol & 84.51 & 84.33 & 85.00 & 84.99 & 85.15 \\
 & Cloze & 60.52 & 68.53 & 70.01 & 71.99 & 72.98 \\
\midrule
CommonsenseQA & Symbol & 66.50 & 66.09 & 66.80 & 66.17 & 65.93 \\
 & Cloze & 44.39 & 58.94 & 64.36 & 67.19 & 69.04 \\
\midrule
MMLU & Symbol & 56.97 & 57.33 & 57.49 & 57.65 & 57.64 \\
 & Cloze & 38.09 & 40.06 & 40.86 & 42.24 & 43.30 \\
\midrule
OpenBookQA & Symbol & 64.80 & 68.40 & 68.84 & 68.88 & 70.68 \\
 & Cloze & 43.20 & 43.68 & 44.00 & 45.52 & 46.72 \\
\midrule
HellaSwag & Symbol & 25.13 & 40.54 & 40.86 & 40.39 & 39.34 \\
 & Cloze & 72.69 & 72.21 & 72.68 & 73.60 & 73.80 \\
\midrule
WinoGrande & Symbol & 50.83 & 52.11 & 52.00 & 51.98 & 52.14 \\
 & Cloze & 66.93 & 65.52 & 66.08 & 66.71 & 66.99 \\
\midrule
PIQA & Symbol & 54.19 & 69.77 & 74.43 & 74.04 & 74.46 \\
 & Cloze & 77.80 & 77.50 & 77.95 & 78.73 & 79.09 \\
\bottomrule
\end{tabular}
\caption{Performance (\%) comparing Symbol and Cloze formats across $k$-shot settings ($k \in \{0,1,2,5,10\}$) for Llama-3.2-3B.}
\label{tab:symbol_cloze_llama_3_2_3b}
\end{table}

\begin{table}[!htbp]
\centering
\small
\setlength{\tabcolsep}{4pt}
\begin{tabular}{p{0.20\textwidth} p{0.11\textwidth} rrrrr}
\toprule
\textbf{Task} & \textbf{Method} & \textbf{0-shot} & \textbf{1-shot} & \textbf{2-shot} & \textbf{5-shot} & \textbf{10-shot} \\
\midrule
ARC-Challenge & Symbol & 73.72 & 74.03 & 74.76 & 74.83 & 74.64 \\
 & Cloze & 39.25 & 40.02 & 41.07 & 41.69 & 44.06 \\
\midrule
ARC-Easy & Symbol & 88.05 & 87.07 & 87.74 & 88.33 & 88.75 \\
 & Cloze & 65.53 & 68.03 & 70.08 & 72.20 & 73.80 \\
\midrule
CommonsenseQA & Symbol & 73.87 & 70.66 & 71.83 & 72.53 & 72.30 \\
 & Cloze & 49.06 & 59.12 & 64.06 & 67.11 & 68.32 \\
\midrule
MMLU & Symbol & 58.62 & 59.49 & 60.80 & 61.64 & 61.80 \\
 & Cloze & 38.42 & 39.44 & 40.11 & 40.92 & 42.13 \\
\midrule
OpenBookQA & Symbol & 76.20 & 75.12 & 75.16 & 76.60 & 77.68 \\
 & Cloze & 39.20 & 39.88 & 41.12 & 41.88 & 41.64 \\
\midrule
HellaSwag & Symbol & 34.49 & 58.52 & 59.48 & 57.75 & 56.77 \\
 & Cloze & 70.00 & 69.88 & 70.14 & 70.95 & 71.02 \\
\midrule
WinoGrande & Symbol & 53.28 & 54.55 & 54.99 & 55.52 & 54.87 \\
 & Cloze & 63.69 & 63.22 & 64.34 & 64.99 & 65.41 \\
\midrule
PIQA & Symbol & 65.18 & 74.45 & 74.81 & 75.64 & 75.65 \\
 & Cloze & 76.82 & 75.74 & 76.14 & 76.90 & 77.10 \\
\bottomrule
\end{tabular}
\caption{Performance (\%) comparing Symbol and Cloze formats across $k$-shot settings ($k \in \{0,1,2,5,10\}$) for Llama-3.2-3B-Instruct.}
\label{tab:symbol_cloze_llama_3_2_3b_inst}
\end{table}

\begin{table}[!htbp]
\centering
\small
\setlength{\tabcolsep}{4pt}
\begin{tabular}{p{0.20\textwidth} p{0.11\textwidth} rrrrr}
\toprule
\textbf{Task} & \textbf{Method} & \textbf{0-shot} & \textbf{1-shot} & \textbf{2-shot} & \textbf{5-shot} & \textbf{10-shot} \\
\midrule
ARC-Challenge & Symbol & 79.44 & 79.59 & 79.61 & 80.14 & 80.26 \\
 & Cloze & 41.89 & 42.81 & 43.26 & 46.49 & 49.88 \\
\midrule
ARC-Easy & Symbol & 91.88 & 91.82 & 91.61 & 91.81 & 91.81 \\
 & Cloze & 63.76 & 70.81 & 73.68 & 76.67 & 79.06 \\
\midrule
CommonsenseQA & Symbol & 70.76 & 73.07 & 72.94 & 74.32 & 75.46 \\
 & Cloze & 45.86 & 65.93 & 67.83 & 70.76 & 72.32 \\
\midrule
MMLU & Symbol & 64.68 & 65.62 & 65.91 & 66.80 & 66.93 \\
 & Cloze & 41.35 & 44.06 & 45.31 & 46.85 & 48.44 \\
\midrule
OpenBookQA & Symbol & 75.60 & 78.64 & 79.32 & 79.36 & 80.00 \\
 & Cloze & 44.80 & 45.72 & 47.00 & 48.92 & 49.20 \\
\midrule
HellaSwag & Symbol & 25.19 & 50.94 & 56.51 & 56.06 & 53.82 \\
 & Cloze & 78.48 & 77.27 & 77.60 & 78.68 & 78.81 \\
\midrule
WinoGrande & Symbol & 52.64 & 54.46 & 57.65 & 59.83 & 60.90 \\
 & Cloze & 70.56 & 70.69 & 72.50 & 73.72 & 73.87 \\
\midrule
PIQA & Symbol & 51.63 & 77.93 & 80.84 & 80.62 & 79.30 \\
 & Cloze & 81.23 & 80.85 & 81.08 & 82.14 & 82.03 \\
\bottomrule
\end{tabular}
\caption{Performance (\%) comparing Symbol and Cloze formats across $k$-shot settings ($k \in \{0,1,2,5,10\}$) for Llama-3.1-8B.}
\label{tab:symbol_cloze_llama_3_1_8b}
\end{table}

\begin{table}[!htbp]
\centering
\small
\setlength{\tabcolsep}{4pt}
\begin{tabular}{p{0.20\textwidth} p{0.11\textwidth} rrrrr}
\toprule
\textbf{Task} & \textbf{Method} & \textbf{0-shot} & \textbf{1-shot} & \textbf{2-shot} & \textbf{5-shot} & \textbf{10-shot} \\
\midrule
ARC-Challenge & Symbol & 82.08 & 82.30 & 83.11 & 83.31 & 83.45 \\
 & Cloze & 43.26 & 44.95 & 47.30 & 52.68 & 56.18 \\
\midrule
ARC-Easy & Symbol & 93.27 & 93.06 & 93.08 & 93.26 & 93.50 \\
 & Cloze & 67.38 & 73.43 & 77.03 & 81.04 & 82.47 \\
\midrule
CommonsenseQA & Symbol & 75.92 & 76.10 & 77.77 & 78.49 & 78.67 \\
 & Cloze & 47.58 & 66.95 & 69.55 & 70.70 & 72.22 \\
\midrule
MMLU & Symbol & 65.41 & 67.45 & 68.46 & 68.95 & 69.26 \\
 & Cloze & 41.88 & 44.09 & 46.35 & 48.46 & 49.70 \\
\midrule
OpenBookQA & Symbol & 81.00 & 81.76 & 82.36 & 83.56 & 83.44 \\
 & Cloze & 46.40 & 45.32 & 47.12 & 48.08 & 48.72 \\
\midrule
HellaSwag & Symbol & 44.90 & 73.51 & 74.29 & 74.24 & 72.92 \\
 & Cloze & 77.21 & 77.26 & 77.31 & 77.89 & 77.99 \\
\midrule
WinoGrande & Symbol & 58.33 & 63.79 & 65.30 & 66.38 & 66.72 \\
 & Cloze & 69.53 & 70.70 & 71.51 & 73.61 & 73.72 \\
\midrule
PIQA & Symbol & 65.61 & 83.36 & 83.47 & 83.17 & 82.92 \\
 & Cloze & 80.25 & 79.87 & 80.51 & 81.13 & 80.69 \\
\bottomrule
\end{tabular}
\caption{Performance (\%) comparing Symbol and Cloze formats across $k$-shot settings ($k \in \{0,1,2,5,10\}$) for Llama-3.1-8B-Instruct.}
\label{tab:symbol_cloze_llama_3_1_8b_inst}
\end{table}

% ======================
% Mistral
% ======================
\begin{table}[!t]
\centering
\small
\setlength{\tabcolsep}{4pt}
\begin{tabular}{p{0.20\textwidth} p{0.11\textwidth} rrrrr}
\toprule
\textbf{Task} & \textbf{Method} & \textbf{0-shot} & \textbf{1-shot} & \textbf{2-shot} & \textbf{5-shot} & \textbf{10-shot} \\
\midrule
ARC-Challenge & Symbol & 76.62 & 75.94 & 77.39 & 78.57 & 78.48 \\
 & Cloze & 43.86 & 45.14 & 47.22 & 49.51 & 52.25 \\
\midrule
ARC-Easy & Symbol & 88.59 & 88.62 & 89.41 & 89.60 & 90.16 \\
 & Cloze & 63.85 & 70.22 & 72.75 & 76.18 & 78.36 \\
\midrule
CommonsenseQA & Symbol & 57.49 & 69.29 & 71.06 & 72.29 & 72.04 \\
 & Cloze & 44.06 & 59.07 & 67.70 & 70.86 & 72.19 \\
\midrule
MMLU & Symbol & 60.02 & 61.88 & 62.33 & 63.19 & 63.54 \\
 & Cloze & 40.76 & 42.97 & 44.46 & 46.01 & 47.41 \\
\midrule
OpenBookQA & Symbol & 68.20 & 77.44 & 78.68 & 79.60 & 79.48 \\
 & Cloze & 46.40 & 46.08 & 46.52 & 48.40 & 48.68 \\
\midrule
HellaSwag & Symbol & 25.28 & 38.32 & 46.52 & 43.77 & 43.39 \\
 & Cloze & 78.56 & 78.76 & 79.06 & 79.78 & 80.01 \\
\midrule
WinoGrande & Symbol & 53.51 & 55.04 & 57.58 & 58.59 & 58.12 \\
 & Cloze & 67.40 & 67.72 & 68.89 & 71.29 & 72.20 \\
\midrule
PIQA & Symbol & 52.77 & 69.96 & 78.01 & 79.14 & 79.14 \\
 & Cloze & 81.23 & 81.06 & 81.48 & 82.11 & 82.41 \\
\bottomrule
\end{tabular}
\caption{Performance (\%) comparing Symbol and Cloze formats across $k$-shot settings ($k \in \{0,1,2,5,10\}$) for Mistral-7B-v0.3.}
\label{tab:symbol_cloze_mistral_7b_v0_3}
\end{table}

\begin{table}[!t]
\centering
\small
\setlength{\tabcolsep}{4pt}
\begin{tabular}{p{0.20\textwidth} p{0.11\textwidth} rrrrr}
\toprule
\textbf{Task} & \textbf{Method} & \textbf{0-shot} & \textbf{1-shot} & \textbf{2-shot} & \textbf{5-shot} & \textbf{10-shot} \\
\midrule
ARC-Challenge & Symbol & 77.90 & 77.53 & 78.23 & 78.75 & 78.81 \\
 & Cloze & 49.15 & 52.39 & 54.69 & 58.80 & 60.75 \\
\midrule
ARC-Easy & Symbol & 88.17 & 89.38 & 89.77 & 90.07 & 90.22 \\
 & Cloze & 69.57 & 76.60 & 79.41 & 82.04 & 83.21 \\
\midrule
CommonsenseQA & Symbol & 69.37 & 72.81 & 73.32 & 73.53 & 74.32 \\
 & Cloze & 47.99 & 66.48 & 69.70 & 72.74 & 73.25 \\
\midrule
MMLU & Symbol & 60.28 & 61.34 & 61.88 & 62.57 & 62.93 \\
 & Cloze & 44.05 & 46.28 & 47.42 & 48.68 & 49.83 \\
\midrule
OpenBookQA & Symbol & 73.60 & 78.12 & 78.28 & 79.96 & 80.80 \\
 & Cloze & 48.60 & 48.28 & 49.12 & 51.84 & 52.36 \\
\midrule
HellaSwag & Symbol & 53.54 & 67.75 & 70.39 & 70.06 & 70.02 \\
 & Cloze & 81.18 & 81.28 & 81.43 & 81.90 & 82.11 \\
\midrule
WinoGrande & Symbol & 58.56 & 58.39 & 60.79 & 62.00 & 60.85 \\
 & Cloze & 69.69 & 70.26 & 71.82 & 73.75 & 74.16 \\
\midrule
PIQA & Symbol & 75.19 & 77.07 & 79.97 & 81.12 & 80.73 \\
 & Cloze & 82.64 & 81.91 & 82.95 & 83.61 & 83.68 \\
\bottomrule
\end{tabular}
\caption{Performance (\%) comparing Symbol and Cloze formats across $k$-shot settings ($k \in \{0,1,2,5,10\}$) for Mistral-7B-Inst-v0.3.}
\label{tab:symbol_cloze_mistral_7b_inst_v0_3}
\end{table}

\begin{table}[!t]
\centering
\small
\setlength{\tabcolsep}{4pt}
\begin{tabular}{p{0.20\textwidth} p{0.11\textwidth} rrrrr}
\toprule
\textbf{Task} & \textbf{Method} & \textbf{0-shot} & \textbf{1-shot} & \textbf{2-shot} & \textbf{5-shot} & \textbf{10-shot} \\
\midrule
ARC-Challenge & Symbol & 92.75 & 93.33 & 93.26 & 93.52 & 93.76 \\
 & Cloze & 48.46 & 53.82 & 53.94 & 58.50 & 62.22 \\
\midrule
ARC-Easy & Symbol & 97.77 & 98.42 & 98.33 & 98.38 & 98.48 \\
 & Cloze & 70.54 & 77.40 & 80.31 & 83.07 & 84.69 \\
\midrule
CommonsenseQA & Symbol & 70.84 & 81.52 & 81.57 & 83.62 & 84.32 \\
 & Cloze & 47.17 & 68.88 & 71.53 & 73.53 & 75.20 \\
\midrule
MMLU & Symbol & 78.23 & 80.20 & 80.44 & 80.65 & 81.14 \\
 & Cloze & 46.81 & 52.17 & 54.33 & 57.03 & 57.84 \\
\midrule
OpenBookQA & Symbol & 85.20 & 90.08 & 91.64 & 93.16 & 93.04 \\
 & Cloze & 51.00 & 49.36 & 51.12 & 53.04 & 52.08 \\
\midrule
HellaSwag & Symbol & 25.22 & 74.61 & 81.40 & 81.55 & 82.00 \\
 & Cloze & 80.99 & 81.12 & 81.67 & 82.54 & 82.80 \\
\midrule
WinoGrande & Symbol & 68.43 & 76.81 & 80.40 & 81.74 & 81.28 \\
 & Cloze & 75.93 & 76.51 & 78.36 & 78.94 & 79.20 \\
\midrule
PIQA & Symbol & 55.93 & 87.44 & 87.80 & 88.85 & 88.84 \\
 & Cloze & 83.19 & 82.70 & 82.84 & 83.32 & 83.60 \\
\bottomrule
\end{tabular}
\caption{Performance (\%) comparing Symbol and Cloze formats across $k$-shot settings ($k \in \{0,1,2,5,10\}$) for Mistral-Small-24B-Base-2501.}
\label{tab:symbol_cloze_mistral_small_24b_base_2501}
\end{table}

\begin{table}[!t]
\centering
\small
\setlength{\tabcolsep}{4pt}
\begin{tabular}{p{0.20\textwidth} p{0.11\textwidth} rrrrr}
\toprule
\textbf{Task} & \textbf{Method} & \textbf{0-shot} & \textbf{1-shot} & \textbf{2-shot} & \textbf{5-shot} & \textbf{10-shot} \\
\midrule
ARC-Challenge & Symbol & 93.94 & 93.77 & 93.72 & 94.05 & 94.21 \\
 & Cloze & 51.19 & 58.75 & 61.93 & 66.88 & 67.85 \\
\midrule
ARC-Easy & Symbol & 98.23 & 98.32 & 98.33 & 98.36 & 98.44 \\
 & Cloze & 73.23 & 83.77 & 86.56 & 87.98 & 88.56 \\
\midrule
CommonsenseQA & Symbol & 82.23 & 83.92 & 84.32 & 84.85 & 85.37 \\
 & Cloze & 54.79 & 72.38 & 74.30 & 74.91 & 76.50 \\
\midrule
MMLU & Symbol & 79.30 & 80.77 & 80.82 & 81.20 & 81.24 \\
 & Cloze & 47.72 & 52.80 & 55.95 & 58.35 & 58.93 \\
\midrule
OpenBookQA & Symbol & 91.80 & 92.92 & 93.68 & 94.48 & 94.48 \\
 & Cloze & 50.40 & 50.36 & 52.40 & 53.88 & 53.32 \\
\midrule
HellaSwag & Symbol & 86.33 & 93.22 & 93.55 & 93.59 & 93.78 \\
 & Cloze & 80.73 & 80.50 & 80.93 & 81.71 & 81.92 \\
\midrule
WinoGrande & Symbol & 76.09 & 79.56 & 81.17 & 82.48 & 83.31 \\
 & Cloze & 74.74 & 76.38 & 78.42 & 79.15 & 79.11 \\
\midrule
PIQA & Symbol & 85.04 & 91.21 & 91.17 & 91.13 & 91.76 \\
 & Cloze & 83.24 & 83.31 & 83.60 & 84.34 & 84.30 \\
\bottomrule
\end{tabular}
\caption{Performance (\%) comparing Symbol and Cloze formats across $k$-shot settings ($k \in \{0,1,2,5,10\}$) for Mistral-Small-24B-Inst-2501.}
\label{tab:symbol_cloze_mistral_small_24b_inst_2501}
\end{table}

\begin{table}[!t]
\centering
\small
\setlength{\tabcolsep}{4pt}
\begin{tabular}{p{0.20\textwidth} p{0.11\textwidth} rrrrr}
\toprule
\textbf{Task} & \textbf{Method} & \textbf{0-shot} & \textbf{1-shot} & \textbf{2-shot} & \textbf{5-shot} & \textbf{10-shot} \\
\midrule
ARC-Challenge & Symbol & 81.06 & 84.01 & 84.80 & 85.14 & 84.90 \\
 & Cloze & 46.84 & 50.26 & 52.44 & 56.52 & 59.20 \\
\midrule
ARC-Easy & Symbol & 91.04 & 93.32 & 93.31 & 93.86 & 94.24 \\
 & Cloze & 67.55 & 77.41 & 79.84 & 82.41 & 83.88 \\
\midrule
CommonsenseQA & Symbol & 55.94 & 72.19 & 74.17 & 75.41 & 76.71 \\
 & Cloze & 47.09 & 62.33 & 69.43 & 72.07 & 74.02 \\
\midrule
MMLU & Symbol & 63.91 & 67.13 & 68.25 & 69.32 & 69.97 \\
 & Cloze & 42.75 & 46.26 & 48.01 & 49.74 & 51.29 \\
\midrule
OpenBookQA & Symbol & 73.20 & 80.00 & 81.20 & 82.96 & 84.00 \\
 & Cloze & 49.40 & 47.52 & 48.24 & 50.76 & 51.32 \\
\midrule
HellaSwag & Symbol & 25.23 & 43.07 & 51.69 & 49.51 & 49.54 \\
 & Cloze & 79.99 & 80.56 & 81.36 & 82.20 & 82.17 \\
\midrule
WinoGrande & Symbol & 55.56 & 59.36 & 63.52 & 68.65 & 70.32 \\
 & Cloze & 72.30 & 73.86 & 76.18 & 78.18 & 78.34 \\
\midrule
PIQA & Symbol & 50.54 & 75.64 & 82.94 & 82.74 & 83.20 \\
 & Cloze & 82.21 & 81.79 & 82.78 & 82.87 & 83.10 \\
\bottomrule
\end{tabular}
\caption{Performance (\%) comparing Symbol and Cloze formats across $k$-shot settings ($k \in \{0,1,2,5,10\}$) for Mistral-Nemo-Base-2407.}
\label{tab:symbol_cloze_mistral_nemo_base_2407}
\end{table}

\begin{table}[!t]
\centering
\small
\setlength{\tabcolsep}{4pt}
\begin{tabular}{p{0.20\textwidth} p{0.11\textwidth} rrrrr}
\toprule
\textbf{Task} & \textbf{Method} & \textbf{0-shot} & \textbf{1-shot} & \textbf{2-shot} & \textbf{5-shot} & \textbf{10-shot} \\
\midrule
ARC-Challenge & Symbol & 74.06 & 83.21 & 84.62 & 84.88 & 85.20 \\
 & Cloze & 47.87 & 51.57 & 54.35 & 58.27 & 59.69 \\
\midrule
ARC-Easy & Symbol & 84.09 & 90.82 & 92.74 & 93.88 & 94.16 \\
 & Cloze & 70.71 & 77.11 & 80.22 & 83.67 & 84.90 \\
\midrule
CommonsenseQA & Symbol & 61.92 & 72.65 & 75.94 & 77.72 & 77.89 \\
 & Cloze & 50.61 & 67.31 & 71.30 & 72.94 & 74.55 \\
\midrule
MMLU & Symbol & 59.17 & 65.30 & 67.16 & 68.95 & 69.38 \\
 & Cloze & 43.91 & 46.74 & 48.34 & 50.54 & 51.79 \\
\midrule
OpenBookQA & Symbol & 65.80 & 77.60 & 81.60 & 84.48 & 85.00 \\
 & Cloze & 51.40 & 48.48 & 50.04 & 52.56 & 51.80 \\
\midrule
HellaSwag & Symbol & 30.96 & 77.13 & 77.61 & 74.38 & 71.37 \\
 & Cloze & 80.25 & 80.07 & 80.44 & 81.61 & 81.80 \\
\midrule
WinoGrande & Symbol & 57.93 & 62.65 & 66.71 & 71.08 & 72.39 \\
 & Cloze & 73.32 & 73.73 & 76.20 & 77.58 & 77.87 \\
\midrule
PIQA & Symbol & 59.68 & 86.09 & 84.56 & 85.14 & 84.84 \\
 & Cloze & 81.66 & 81.66 & 82.53 & 82.89 & 82.91 \\
\bottomrule
\end{tabular}
\caption{Performance (\%) comparing Symbol and Cloze formats across $k$-shot settings ($k \in \{0,1,2,5,10\}$) for Mistral-Nemo-Inst-2407.}
\label{tab:symbol_cloze_mistral_nemo_inst_2407}
\end{table}

% ======================
% Qwen3
% ======================
\begin{table}[!t]
\centering
\small
\setlength{\tabcolsep}{4pt}
\begin{tabular}{p{0.20\textwidth} p{0.11\textwidth} rrrrr}
\toprule
\textbf{Task} & \textbf{Method} & \textbf{0-shot} & \textbf{1-shot} & \textbf{2-shot} & \textbf{5-shot} & \textbf{10-shot} \\
\midrule
ARC-Challenge & Symbol & 62.80 & 66.95 & 67.03 & 66.33 & 66.76 \\
 & Cloze & 33.70 & 36.79 & 37.56 & 40.68 & 42.18 \\
\midrule
ARC-Easy & Symbol & 81.36 & 82.55 & 83.24 & 83.44 & 83.81 \\
 & Cloze & 52.40 & 63.67 & 67.60 & 70.72 & 71.14 \\
\midrule
CommonsenseQA & Symbol & 58.56 & 61.72 & 62.15 & 62.60 & 62.85 \\
 & Cloze & 27.44 & 51.19 & 53.94 & 56.04 & 57.66 \\
\midrule
MMLU & Symbol & 51.34 & 53.45 & 54.19 & 54.44 & 54.65 \\
 & Cloze & 33.55 & 35.69 & 36.89 & 38.07 & 38.80 \\
\midrule
OpenBookQA & Symbol & 58.00 & 61.64 & 62.60 & 63.88 & 64.56 \\
 & Cloze & 36.00 & 36.56 & 36.92 & 38.80 & 39.24 \\
\midrule
HellaSwag & Symbol & 29.72 & 43.56 & 44.45 & 44.13 & 42.67 \\
 & Cloze & 50.68 & 50.98 & 51.00 & 51.31 & 51.38 \\
\midrule
WinoGrande & Symbol & 50.67 & 53.48 & 52.88 & 51.86 & 51.96 \\
 & Cloze & 55.72 & 56.09 & 56.19 & 56.15 & 56.39 \\
\midrule
PIQA & Symbol & 60.23 & 65.00 & 67.28 & 64.78 & 65.31 \\
 & Cloze & 70.89 & 70.13 & 70.23 & 70.61 & 70.46 \\
\bottomrule
\end{tabular}
\caption{Performance (\%) comparing Symbol and Cloze formats across $k$-shot settings ($k \in \{0,1,2,5,10\}$) for Qwen3-0.6B-Base.}
\label{tab:symbol_cloze_qwen3_0_6b_base}
\end{table}

\begin{table}[!t]
\centering
\small
\setlength{\tabcolsep}{4pt}
\begin{tabular}{p{0.20\textwidth} p{0.11\textwidth} rrrrr}
\toprule
\textbf{Task} & \textbf{Method} & \textbf{0-shot} & \textbf{1-shot} & \textbf{2-shot} & \textbf{5-shot} & \textbf{10-shot} \\
\midrule
ARC-Challenge & Symbol & 52.05 & 50.68 & 57.37 & 59.47 & 59.49 \\
 & Cloze & 30.38 & 31.06 & 31.74 & 34.13 & 35.89 \\
\midrule
ARC-Easy & Symbol & 71.63 & 69.60 & 75.65 & 76.88 & 77.52 \\
 & Cloze & 49.66 & 53.58 & 57.63 & 63.87 & 65.83 \\
\midrule
CommonsenseQA & Symbol & 46.68 & 50.38 & 52.38 & 52.92 & 53.99 \\
 & Cloze & 28.09 & 34.33 & 44.06 & 46.65 & 51.63 \\
\midrule
MMLU & Symbol & 43.04 & 43.44 & 46.55 & 48.28 & 49.49 \\
 & Cloze & 31.03 & 31.70 & 32.50 & 33.69 & 34.19 \\
\midrule
OpenBookQA & Symbol & 52.20 & 47.32 & 44.84 & 54.04 & 55.80 \\
 & Cloze & 35.00 & 35.76 & 36.24 & 37.60 & 39.48 \\
\midrule
HellaSwag & Symbol & 25.76 & 36.02 & 38.43 & 40.71 & 41.79 \\
 & Cloze & 45.05 & 44.64 & 44.31 & 44.43 & 44.51 \\
\midrule
WinoGrande & Symbol & 49.80 & 51.03 & 52.75 & 52.36 & 51.54 \\
 & Cloze & 52.49 & 52.44 & 51.77 & 52.77 & 53.13 \\
\midrule
PIQA & Symbol & 50.82 & 58.30 & 59.28 & 60.77 & 61.14 \\
 & Cloze & 65.72 & 66.60 & 66.53 & 67.04 & 67.00 \\
\bottomrule
\end{tabular}
\caption{Performance (\%) comparing Symbol and Cloze formats across $k$-shot settings ($k \in \{0,1,2,5,10\}$) for Qwen3-0.6B.}
\label{tab:symbol_cloze_qwen3_0_6b}
\end{table}

\begin{table}[!t]
\centering
\small
\setlength{\tabcolsep}{4pt}
\begin{tabular}{p{0.20\textwidth} p{0.11\textwidth} rrrrr}
\toprule
\textbf{Task} & \textbf{Method} & \textbf{0-shot} & \textbf{1-shot} & \textbf{2-shot} & \textbf{5-shot} & \textbf{10-shot} \\
\midrule
ARC-Challenge & Symbol & 80.29 & 81.11 & 81.45 & 81.79 & 82.00 \\
 & Cloze & 39.76 & 45.92 & 49.27 & 54.47 & 54.85 \\
\midrule
ARC-Easy & Symbol & 91.75 & 91.78 & 92.25 & 92.62 & 92.84 \\
 & Cloze & 60.06 & 73.63 & 78.00 & 79.56 & 79.88 \\
\midrule
CommonsenseQA & Symbol & 72.48 & 73.66 & 73.77 & 74.50 & 74.48 \\
 & Cloze & 31.78 & 61.90 & 64.80 & 66.14 & 67.27 \\
\midrule
MMLU & Symbol & 62.90 & 63.85 & 64.41 & 64.94 & 65.36 \\
 & Cloze & 37.09 & 42.11 & 44.03 & 45.22 & 45.85 \\
\midrule
OpenBookQA & Symbol & 72.00 & 76.44 & 77.36 & 77.92 & 79.24 \\
 & Cloze & 42.60 & 40.76 & 43.28 & 45.00 & 45.96 \\
\midrule
HellaSwag & Symbol & 29.02 & 61.60 & 60.53 & 60.57 & 58.60 \\
 & Cloze & 63.17 & 63.35 & 63.62 & 63.84 & 64.05 \\
\midrule
WinoGrande & Symbol & 51.85 & 54.06 & 55.52 & 56.16 & 56.20 \\
 & Cloze & 60.69 & 59.62 & 60.05 & 61.37 & 61.54 \\
\midrule
PIQA & Symbol & 57.29 & 77.77 & 77.22 & 76.94 & 77.53 \\
 & Cloze & 74.86 & 75.12 & 75.67 & 76.43 & 76.42 \\
\bottomrule
\end{tabular}
\caption{Performance (\%) comparing Symbol and Cloze formats across $k$-shot settings ($k \in \{0,1,2,5,10\}$) for Qwen3-1.7B-Base.}
\label{tab:symbol_cloze_qwen3_1_7b_base}
\end{table}

\begin{table}[!t]
\centering
\small
\setlength{\tabcolsep}{4pt}
\begin{tabular}{p{0.20\textwidth} p{0.11\textwidth} rrrrr}
\toprule
\textbf{Task} & \textbf{Method} & \textbf{0-shot} & \textbf{1-shot} & \textbf{2-shot} & \textbf{5-shot} & \textbf{10-shot} \\
\midrule
ARC-Challenge & Symbol & 73.89 & 76.04 & 76.61 & 77.56 & 77.90 \\
 & Cloze & 35.32 & 40.75 & 44.23 & 50.34 & 51.52 \\
\midrule
ARC-Easy & Symbol & 87.84 & 88.92 & 89.13 & 89.91 & 90.03 \\
 & Cloze & 54.84 & 67.85 & 74.15 & 77.24 & 78.13 \\
\midrule
CommonsenseQA & Symbol & 64.29 & 68.27 & 68.60 & 69.71 & 70.27 \\
 & Cloze & 29.32 & 50.60 & 60.41 & 63.75 & 65.05 \\
\midrule
MMLU & Symbol & 57.14 & 59.83 & 60.64 & 61.87 & 62.07 \\
 & Cloze & 35.02 & 37.44 & 40.14 & 42.52 & 43.43 \\
\midrule
OpenBookQA & Symbol & 65.80 & 69.96 & 69.24 & 71.72 & 72.84 \\
 & Cloze & 41.00 & 40.96 & 42.12 & 45.00 & 45.44 \\
\midrule
HellaSwag & Symbol & 25.07 & 53.07 & 56.13 & 59.50 & 60.99 \\
 & Cloze & 57.57 & 57.40 & 57.57 & 57.73 & 57.77 \\
\midrule
WinoGrande & Symbol & 52.01 & 51.84 & 54.62 & 55.12 & 55.11 \\
 & Cloze & 57.46 & 56.48 & 56.73 & 57.11 & 57.40 \\
\midrule
PIQA & Symbol & 55.01 & 68.26 & 71.56 & 73.42 & 73.08 \\
 & Cloze & 72.58 & 72.36 & 72.77 & 74.16 & 74.36 \\
\bottomrule
\end{tabular}
\caption{Performance (\%) comparing Symbol and Cloze formats across $k$-shot settings ($k \in \{0,1,2,5,10\}$) for Qwen3-1.7B.}
\label{tab:symbol_cloze_qwen3_1_7b}
\end{table}

\begin{table}[!t]
\centering
\small
\setlength{\tabcolsep}{4pt}
\begin{tabular}{p{0.20\textwidth} p{0.11\textwidth} rrrrr}
\toprule
\textbf{Task} & \textbf{Method} & \textbf{0-shot} & \textbf{1-shot} & \textbf{2-shot} & \textbf{5-shot} & \textbf{10-shot} \\
\midrule
ARC-Challenge & Symbol & 88.05 & 89.52 & 89.35 & 89.54 & 89.93 \\
 & Cloze & 42.15 & 51.52 & 55.17 & 60.29 & 61.29 \\
\midrule
ARC-Easy & Symbol & 96.42 & 96.86 & 96.98 & 96.92 & 96.95 \\
 & Cloze & 61.62 & 77.31 & 82.34 & 84.02 & 84.81 \\
\midrule
CommonsenseQA & Symbol & 82.47 & 80.56 & 81.03 & 81.59 & 81.92 \\
 & Cloze & 30.38 & 67.96 & 71.30 & 71.65 & 72.43 \\
\midrule
MMLU & Symbol & 72.63 & 74.50 & 75.14 & 75.49 & 75.67 \\
 & Cloze & 40.70 & 47.93 & 50.19 & 51.72 & 52.54 \\
\midrule
OpenBookQA & Symbol & 81.40 & 84.76 & 85.92 & 87.28 & 87.40 \\
 & Cloze & 42.60 & 43.76 & 45.76 & 49.44 & 49.72 \\
\midrule
HellaSwag & Symbol & 57.12 & 77.97 & 81.54 & 81.68 & 82.18 \\
 & Cloze & 70.75 & 71.80 & 72.20 & 72.51 & 72.64 \\
\midrule
WinoGrande & Symbol & 59.75 & 65.59 & 67.20 & 67.80 & 68.78 \\
 & Cloze & 65.75 & 64.18 & 65.81 & 67.02 & 67.78 \\
\midrule
PIQA & Symbol & 80.09 & 84.59 & 84.44 & 85.37 & 85.61 \\
 & Cloze & 77.75 & 77.49 & 77.80 & 78.49 & 78.40 \\
\bottomrule
\end{tabular}
\caption{Performance (\%) comparing Symbol and Cloze formats across $k$-shot settings ($k \in \{0,1,2,5,10\}$) for Qwen3-4B-Base.}
\label{tab:symbol_cloze_qwen3_4b_base}
\end{table}

\begin{table}[!t]
\centering
\small
\setlength{\tabcolsep}{4pt}
\begin{tabular}{p{0.20\textwidth} p{0.11\textwidth} rrrrr}
\toprule
\textbf{Task} & \textbf{Method} & \textbf{0-shot} & \textbf{1-shot} & \textbf{2-shot} & \textbf{5-shot} & \textbf{10-shot} \\
\midrule
ARC-Challenge & Symbol & 87.12 & 87.61 & 88.33 & 88.48 & 88.65 \\
 & Cloze & 41.04 & 48.79 & 52.08 & 57.24 & 58.68 \\
\midrule
ARC-Easy & Symbol & 94.07 & 95.02 & 95.06 & 95.18 & 95.38 \\
 & Cloze & 60.35 & 75.07 & 79.69 & 83.13 & 84.11 \\
\midrule
CommonsenseQA & Symbol & 75.68 & 78.02 & 78.72 & 79.06 & 79.59 \\
 & Cloze & 34.23 & 63.18 & 66.50 & 68.76 & 70.04 \\
\midrule
MMLU & Symbol & 69.51 & 71.24 & 71.51 & 71.85 & 72.29 \\
 & Cloze & 39.71 & 43.82 & 47.06 & 49.69 & 50.72 \\
\midrule
OpenBookQA & Symbol & 78.00 & 81.20 & 81.40 & 83.32 & 83.28 \\
 & Cloze & 40.00 & 41.88 & 44.32 & 46.92 & 48.76 \\
\midrule
HellaSwag & Symbol & 46.25 & 77.60 & 78.15 & 78.07 & 79.31 \\
 & Cloze & 64.83 & 64.82 & 65.46 & 65.96 & 66.45 \\
\midrule
WinoGrande & Symbol & 60.14 & 62.68 & 64.45 & 65.84 & 66.00 \\
 & Cloze & 61.33 & 60.82 & 61.58 & 62.18 & 61.52 \\
\midrule
PIQA & Symbol & 59.85 & 81.72 & 81.79 & 81.71 & 81.31 \\
 & Cloze & 75.03 & 75.21 & 75.98 & 76.89 & 76.72 \\
\bottomrule
\end{tabular}
\caption{Performance (\%) comparing Symbol and Cloze formats across $k$-shot settings ($k \in \{0,1,2,5,10\}$) for Qwen3-4B.}
\label{tab:symbol_cloze_qwen3_4b}
\end{table}

\begin{table}[!t]
\centering
\small
\setlength{\tabcolsep}{4pt}
\begin{tabular}{p{0.20\textwidth} p{0.11\textwidth} rrrrr}
\toprule
\textbf{Task} & \textbf{Method} & \textbf{0-shot} & \textbf{1-shot} & \textbf{2-shot} & \textbf{5-shot} & \textbf{10-shot} \\
\midrule
ARC-Challenge & Symbol & 92.49 & 93.00 & 93.00 & 93.09 & 92.92 \\
 & Cloze & 41.04 & 55.49 & 59.59 & 63.77 & 64.32 \\
\midrule
ARC-Easy & Symbol & 97.22 & 97.52 & 97.65 & 97.77 & 97.87 \\
 & Cloze & 62.46 & 79.70 & 83.84 & 86.27 & 86.61 \\
\midrule
CommonsenseQA & Symbol & 85.91 & 84.77 & 84.60 & 85.70 & 85.62 \\
 & Cloze & 31.12 & 70.83 & 72.47 & 72.51 & 73.76 \\
\midrule
MMLU & Symbol & 75.98 & 77.67 & 78.15 & 78.56 & 78.81 \\
 & Cloze & 42.53 & 51.32 & 54.21 & 56.06 & 56.49 \\
\midrule
OpenBookQA & Symbol & 85.00 & 87.92 & 88.96 & 89.32 & 89.40 \\
 & Cloze & 43.80 & 45.16 & 47.96 & 51.68 & 51.48 \\
\midrule
HellaSwag & Symbol & 29.15 & 84.48 & 85.50 & 85.24 & 85.45 \\
 & Cloze & 75.77 & 76.42 & 76.54 & 76.52 & 76.72 \\
\midrule
WinoGrande & Symbol & 64.17 & 68.37 & 70.21 & 71.67 & 72.03 \\
 & Cloze & 67.96 & 68.59 & 70.62 & 71.89 & 72.12 \\
\midrule
PIQA & Symbol & 62.84 & 87.42 & 87.58 & 88.68 & 88.89 \\
 & Cloze & 79.43 & 79.09 & 79.79 & 80.32 & 80.31 \\
\bottomrule
\end{tabular}
\caption{Performance (\%) comparing Symbol and Cloze formats across $k$-shot settings ($k \in \{0,1,2,5,10\}$) for Qwen3-8B-Base.}
\label{tab:symbol_cloze_qwen3_8b_base}
\end{table}

\begin{table}[!t]
\centering
\small
\setlength{\tabcolsep}{4pt}
\begin{tabular}{p{0.20\textwidth} p{0.11\textwidth} rrrrr}
\toprule
\textbf{Task} & \textbf{Method} & \textbf{0-shot} & \textbf{1-shot} & \textbf{2-shot} & \textbf{5-shot} & \textbf{10-shot} \\
\midrule
ARC-Challenge & Symbol & 91.21 & 91.35 & 91.62 & 91.60 & 91.84 \\
 & Cloze & 43.26 & 53.02 & 58.60 & 64.04 & 65.37 \\
\midrule
ARC-Easy & Symbol & 96.13 & 96.77 & 96.95 & 97.16 & 97.27 \\
 & Cloze & 61.24 & 78.74 & 84.08 & 86.36 & 86.82 \\
\midrule
CommonsenseQA & Symbol & 78.54 & 81.62 & 82.46 & 82.49 & 83.24 \\
 & Cloze & 32.43 & 67.27 & 69.58 & 72.63 & 74.20 \\
\midrule
MMLU & Symbol & 74.12 & 76.14 & 76.56 & 76.71 & 77.32 \\
 & Cloze & 41.37 & 48.96 & 52.59 & 54.93 & 55.69 \\
\midrule
OpenBookQA & Symbol & 85.00 & 85.44 & 86.24 & 86.72 & 87.32 \\
 & Cloze & 43.80 & 45.08 & 48.24 & 51.44 & 52.56 \\
\midrule
HellaSwag & Symbol & 25.87 & 82.19 & 83.10 & 83.96 & 84.80 \\
 & Cloze & 71.61 & 71.03 & 71.72 & 72.68 & 73.19 \\
\midrule
WinoGrande & Symbol & 65.51 & 67.88 & 68.35 & 70.28 & 69.88 \\
 & Cloze & 64.33 & 64.53 & 65.56 & 66.50 & 66.80 \\
\midrule
PIQA & Symbol & 49.56 & 84.82 & 85.70 & 86.81 & 86.56 \\
 & Cloze & 78.45 & 77.74 & 78.62 & 79.26 & 79.43 \\
\bottomrule
\end{tabular}
\caption{Performance (\%) comparing Symbol and Cloze formats across $k$-shot settings ($k \in \{0,1,2,5,10\}$) for Qwen3-8B.}
\label{tab:symbol_cloze_qwen3_8b}
\end{table}

\begin{table}[!t]
\centering
\small
\setlength{\tabcolsep}{4pt}
\begin{tabular}{p{0.20\textwidth} p{0.11\textwidth} rrrrr}
\toprule
\textbf{Task} & \textbf{Method} & \textbf{0-shot} & \textbf{1-shot} & \textbf{2-shot} & \textbf{5-shot} & \textbf{10-shot} \\
\midrule
ARC-Challenge & Symbol & 93.60 & 93.57 & 93.57 & 93.60 & 93.77 \\
 & Cloze & 42.06 & 57.48 & 60.77 & 66.04 & 66.72 \\
\midrule
ARC-Easy & Symbol & 97.98 & 98.19 & 98.24 & 98.33 & 98.31 \\
 & Cloze & 62.16 & 80.65 & 84.50 & 86.75 & 87.45 \\
\midrule
CommonsenseQA & Symbol & 86.73 & 85.73 & 86.03 & 86.27 & 86.39 \\
 & Cloze & 30.71 & 70.56 & 74.09 & 74.97 & 76.09 \\
\midrule
MMLU & Symbol & 79.46 & 81.50 & 82.06 & 82.58 & 82.55 \\
 & Cloze & 43.32 & 54.15 & 57.12 & 58.67 & 58.88 \\
\midrule
OpenBookQA & Symbol & 90.40 & 91.92 & 92.88 & 93.48 & 93.44 \\
 & Cloze & 45.80 & 47.20 & 49.24 & 53.80 & 52.64 \\
\midrule
HellaSwag & Symbol & 36.07 & 89.87 & 89.65 & 88.86 & 89.05 \\
 & Cloze & 78.86 & 79.45 & 79.59 & 79.97 & 79.84 \\
\midrule
WinoGrande & Symbol & 68.11 & 74.13 & 75.91 & 77.24 & 77.17 \\
 & Cloze & 69.30 & 73.10 & 75.75 & 76.70 & 76.56 \\
\midrule
PIQA & Symbol & 75.52 & 89.56 & 90.02 & 89.98 & 90.13 \\
 & Cloze & 80.63 & 80.90 & 81.31 & 82.22 & 82.47 \\
\bottomrule
\end{tabular}
\caption{Performance (\%) comparing Symbol and Cloze formats across $k$-shot settings ($k \in \{0,1,2,5,10\}$) for Qwen3-14B-Base.}
\label{tab:symbol_cloze_qwen3_14b_base}
\end{table}

\begin{table}[!t]
\centering
\small
\setlength{\tabcolsep}{4pt}
\begin{tabular}{p{0.20\textwidth} p{0.11\textwidth} rrrrr}
\toprule
\textbf{Task} & \textbf{Method} & \textbf{0-shot} & \textbf{1-shot} & \textbf{2-shot} & \textbf{5-shot} & \textbf{10-shot} \\
\midrule
ARC-Challenge & Symbol & 92.32 & 93.84 & 93.65 & 93.59 & 93.45 \\
 & Cloze & 42.75 & 57.25 & 62.08 & 67.10 & 67.98 \\
\midrule
ARC-Easy & Symbol & 97.47 & 97.78 & 97.99 & 98.01 & 98.15 \\
 & Cloze & 62.54 & 81.37 & 85.79 & 87.32 & 87.71 \\
\midrule
CommonsenseQA & Symbol & 80.10 & 83.18 & 84.06 & 84.75 & 84.88 \\
 & Cloze & 36.20 & 70.60 & 72.83 & 74.68 & 75.66 \\
\midrule
MMLU & Symbol & 78.10 & 79.77 & 80.40 & 80.71 & 81.04 \\
 & Cloze & 43.31 & 52.37 & 55.88 & 58.16 & 58.85 \\
\midrule
OpenBookQA & Symbol & 90.00 & 90.68 & 91.20 & 92.32 & 92.80 \\
 & Cloze & 47.20 & 47.64 & 51.24 & 54.08 & 55.44 \\
\midrule
HellaSwag & Symbol & 26.89 & 87.03 & 87.69 & 88.46 & 89.12 \\
 & Cloze & 76.34 & 75.63 & 76.13 & 76.96 & 77.07 \\
\midrule
WinoGrande & Symbol & 71.35 & 73.62 & 75.39 & 76.12 & 75.93 \\
 & Cloze & 67.40 & 69.52 & 69.45 & 71.05 & 71.79 \\
\midrule
PIQA & Symbol & 58.22 & 85.32 & 86.34 & 87.92 & 87.87 \\
 & Cloze & 80.30 & 79.15 & 79.66 & 80.50 & 80.44 \\
\bottomrule
\end{tabular}
\caption{Performance (\%) comparing Symbol and Cloze formats across $k$-shot settings ($k \in \{0,1,2,5,10\}$) for Qwen3-14B.}
\label{tab:symbol_cloze_qwen3_14b}
\end{table}

\FloatBarrier

\FloatBarrier

\section{Evaluation Results with Human-annotated Classifier}
In this section, we present the complete set of results introduced in Section~\ref{subsec:human_eval} by reporting the performance of all models when using a classifier trained on human-annotated labels. The classifier predicts the suitable evaluation format (i.e., symbol or cloze) and the evaluation is then conducted following the predicted format.
Table~\ref{tab:rule_based_router_full} presents the results for all models. We observe performance degradation across almost all models and tasks, suggesting that identifying format mismatches is highly nuanced even for human annotators.

\label{app:human_label}

\FloatBarrier

% tables/app_rule_based_full.tex
\newcommand{\onescore}[2]{\makecell[c]{#1\\(#2)}}

\begin{table}[H]
\centering
\footnotesize
\setlength{\tabcolsep}{3pt}
\resizebox{\textwidth}{!}{%
\begin{tabular}{lccccccccc}
\toprule
\textbf{Model} &
\makecell{\textbf{ARC-}\\\textbf{Challenge}} &
\makecell{\textbf{ARC-}\\\textbf{Easy}} &
\makecell{\textbf{Commonsense-}\\\textbf{QA}} &
\textbf{MMLU} &
\makecell{\textbf{OpenBook-}\\\textbf{QA}} &
\textbf{HellaSwag} &
\makecell{\textbf{Wino-}\\\textbf{Grande}} &
\textbf{PIQA} &
\makecell{\textbf{SocialI-}\\\textbf{QA}}
\\
\midrule
\multicolumn{10}{l}{\emph{Llama\cite{grattafiori2024llama3herdmodels}}} \\
Llama 3.2 1B &
\onescore{44.5}{+0.3} & \onescore{69.1}{+1.7} & \onescore{51.2}{-0.5} & \onescore{39.6}{-2.7} & \onescore{42.4}{-1.2} & \onescore{62.7}{-0.1} & \onescore{58.5}{-1.2} & \onescore{67.1}{-9.6} & \onescore{48.1}{+0.1} \\
Llama 3.2 1B Inst &
\onescore{51.9}{-2.7} & \onescore{73.8}{-1.5} & \onescore{58.9}{-0.2} & \onescore{42.6}{-3.2} & \onescore{46.6}{-13.2} & \onescore{59.9}{+0.0} & \onescore{58.2}{+0.1} & \onescore{67.1}{-7.1} & \onescore{55.0}{+0.0} \\
Llama 3.2 3B &
\onescore{65.2}{-2.6} & \onescore{83.8}{-0.8} & \onescore{66.3}{-0.2} & \onescore{50.3}{-6.6} & \onescore{51.0}{-13.8} & \onescore{72.6}{-0.1} & \onescore{67.6}{-0.7} & \onescore{68.8}{-9.0} & \onescore{59.9}{+0.1} \\
Llama 3.2 3B Inst &
\onescore{70.5}{-3.2} & \onescore{86.2}{-1.8} & \onescore{73.4}{-0.5} & \onescore{52.0}{-6.6} & \onescore{54.8}{-21.4} & \onescore{70.1}{+0.1} & \onescore{63.5}{-0.6} & \onescore{73.6}{-3.3} & \onescore{68.0}{+0.3} \\
Llama 3.1 8B &
\onescore{76.7}{-2.7} & \onescore{90.5}{-1.3} & \onescore{70.4}{-0.3} & \onescore{57.4}{-7.3} & \onescore{58.2}{-17.4} & \onescore{78.4}{-0.1} & \onescore{70.3}{-0.2} & \onescore{70.8}{-10.4} & \onescore{62.7}{-0.1} \\
Llama 3.1 8B Inst &
\onescore{77.6}{-4.4} & \onescore{91.5}{-1.7} & \onescore{75.5}{-0.4} & \onescore{58.4}{-7.0} & \onescore{60.0}{-21.0} & \onescore{77.2}{+0.0} & \onescore{69.7}{-0.9} & \onescore{75.2}{-5.0} & \onescore{70.6}{-0.2} \\
\midrule
\multicolumn{10}{l}{\emph{Mistral\cite{jiang2023mistral7b}}} \\
Mistral 7B v0.3 &
\onescore{74.4}{-2.2} & \onescore{87.2}{-1.4} & \onescore{57.0}{-0.5} & \onescore{53.6}{-6.4} & \onescore{55.8}{-12.4} & \onescore{78.6}{+0.0} & \onescore{67.4}{-4.2} & \onescore{70.8}{-10.4} & \onescore{61.1}{-0.1} \\
Mistral 7B Inst v0.3 &
\onescore{75.8}{-2.1} & \onescore{87.1}{-1.1} & \onescore{69.2}{-0.2} & \onescore{54.5}{-5.8} & \onescore{59.8}{-13.8} & \onescore{81.2}{+0.0} & \onescore{69.5}{-5.4} & \onescore{81.5}{-1.1} & \onescore{68.9}{+0.3} \\
Mistral Small 24B Base 2501 &
\onescore{88.3}{-4.4} & \onescore{96.0}{-1.8} & \onescore{70.5}{-0.3} & \onescore{67.3}{-11.0} & \onescore{63.4}{-21.8} & \onescore{81.0}{+0.0} & \onescore{75.5}{-1.1} & \onescore{73.6}{-9.6} & \onescore{70.1}{+0.1} \\
Mistral Small 24B Inst 2501 &
\onescore{89.3}{-4.6} & \onescore{96.5}{-1.8} & \onescore{82.0}{-0.2} & \onescore{69.1}{-10.2} & \onescore{65.8}{-26.0} & \onescore{80.6}{-5.7} & \onescore{75.3}{-0.9} & \onescore{85.1}{+0.1} & \onescore{79.2}{+0.2} \\
Mistral Nemo Base 2407 &
\onescore{77.0}{-4.1} & \onescore{90.1}{-0.9} & \onescore{55.6}{-2.0} & \onescore{57.0}{-6.9} & \onescore{57.8}{-15.4} & \onescore{80.0}{+0.0} & \onescore{72.1}{-1.5} & \onescore{70.2}{-12.0} & \onescore{64.9}{+0.1} \\
Mistral Nemo Inst 2407 &
\onescore{71.9}{-2.1} & \onescore{84.0}{+0.0} & \onescore{61.7}{-0.2} & \onescore{54.7}{-4.4} & \onescore{55.8}{-10.0} & \onescore{80.2}{+0.0} & \onescore{73.2}{-1.3} & \onescore{75.7}{-5.9} & \onescore{73.4}{+0.0} \\
\midrule
\multicolumn{10}{l}{\emph{Qwen3\cite{yang2025qwen3}}} \\
Qwen3 0.6B Base &
\onescore{59.5}{-3.3} & \onescore{79.3}{-2.0} & \onescore{58.1}{-0.5} & \onescore{44.1}{-7.3} & \onescore{44.6}{-13.4} & \onescore{50.7}{+0.0} & \onescore{55.7}{-0.8} & \onescore{66.2}{-4.9} & \onescore{57.9}{-0.7} \\
Qwen3 0.6B &
\onescore{49.0}{-3.1} & \onescore{71.2}{-0.4} & \onescore{46.3}{-0.4} & \onescore{37.8}{-5.3} & \onescore{42.0}{-10.2} & \onescore{45.1}{+0.0} & \onescore{52.5}{-0.4} & \onescore{60.8}{-5.9} & \onescore{52.0}{-0.3} \\
Qwen3 1.7B Base &
\onescore{76.0}{-4.3} & \onescore{89.4}{-2.3} & \onescore{71.8}{-0.7} & \onescore{52.8}{-10.1} & \onescore{54.6}{-17.4} & \onescore{63.2}{+0.0} & \onescore{60.7}{-0.4} & \onescore{69.9}{-5.0} & \onescore{69.4}{+0.4} \\
Qwen3 1.7B &
\onescore{70.1}{-3.8} & \onescore{86.0}{-1.8} & \onescore{63.6}{-0.7} & \onescore{48.8}{-8.4} & \onescore{51.2}{-14.6} & \onescore{57.6}{+0.0} & \onescore{57.5}{-0.6} & \onescore{65.2}{-7.4} & \onescore{62.5}{+0.2} \\
Qwen3 4B Base &
\onescore{83.3}{-4.8} & \onescore{94.4}{-2.1} & \onescore{81.9}{-0.6} & \onescore{60.6}{-12.0} & \onescore{59.2}{-22.2} & \onescore{70.8}{+0.0} & \onescore{65.7}{-0.3} & \onescore{79.5}{-0.5} & \onescore{74.0}{+0.2} \\
Qwen3 4B &
\onescore{82.3}{-4.9} & \onescore{92.2}{-1.9} & \onescore{75.1}{-0.6} & \onescore{58.3}{-11.3} & \onescore{56.4}{-21.6} & \onescore{64.8}{+0.0} & \onescore{61.3}{+0.0} & \onescore{70.3}{-4.7} & \onescore{72.2}{-0.1} \\
Qwen3 8B Base &
\onescore{87.3}{-5.2} & \onescore{94.8}{-2.4} & \onescore{85.7}{-0.2} & \onescore{63.5}{-12.5} & \onescore{61.0}{-24.0} & \onescore{75.6}{-0.2} & \onescore{68.4}{-0.1} & \onescore{75.3}{-4.1} & \onescore{78.8}{+0.2} \\
Qwen3 8B &
\onescore{86.5}{-4.7} & \onescore{93.9}{-2.2} & \onescore{78.2}{-0.3} & \onescore{62.3}{-11.8} & \onescore{62.0}{-23.0} & \onescore{71.6}{+0.0} & \onescore{64.7}{-0.8} & \onescore{69.0}{-9.4} & \onescore{75.8}{+0.2} \\
Qwen3 14B Base &
\onescore{88.7}{-4.9} & \onescore{95.5}{-2.5} & \onescore{86.6}{-0.2} & \onescore{65.9}{-13.5} & \onescore{66.2}{-24.2} & \onescore{78.8}{-0.1} & \onescore{69.7}{-0.6} & \onescore{78.8}{-1.8} & \onescore{79.8}{-0.1} \\
Qwen3 14B &
\onescore{88.7}{-3.6} & \onescore{95.0}{-2.4} & \onescore{79.5}{-0.6} & \onescore{66.2}{-11.9} & \onescore{63.8}{-26.2} & \onescore{76.4}{+0.0} & \onescore{67.4}{-3.9} & \onescore{73.9}{-6.4} & \onescore{75.9}{+0.1} \\
\bottomrule
\end{tabular}%
}
\caption{Model performance using the classifier trained on human-annotated labels. Values in parentheses indicate the performance gain relative to the baseline with the preferred format ($\max(\text{baseline}_{\texttt{symbol}}, \text{baseline}_{\texttt{cloze}})$).}
\label{tab:rule_based_router_full}
\end{table}

\FloatBarrier

\clearpage

\section{Evaluation Results with Model-generated Classifier}
In this section, we present the complete set of results introduced in Section~\ref{subsec:model_eval} by reporting the performance of all models when using a classifier trained on model-generated labels. For labeling in Algorithm~\ref{alg:per_model_label}, we use a margin threshold of $\delta$ = 0.2. We report model performance with classifiers trained on labels derived from each model itself (self-labeled) as well as from three mid-sized models (majority-labeled).
Table~\ref{tab:model_leveraged_full_llama_mistral} reports the results for the Llama and Mistral model families, while Table~\ref{tab:model_leveraged_full_qwen} presents the results for the Qwen~3 models.

\label{app:model_leveraged}
% tables/app_model_leveraged_full_reformat.tex
% Requires: \usepackage{booktabs} \usepackage{multirow} \usepackage{makecell} \usepackage{adjustbox}

%\newcommand{\onescore}[2]{#1\;(#2)}
%\newcommand{\onescore}[2]{\makecell[c]{#1\\(#2)}}

\begin{table}[H]
\centering
\footnotesize
\setlength{\tabcolsep}{3pt}
\renewcommand{\arraystretch}{1.05}

\begin{adjustbox}{max width=\textwidth}
\begin{tabular}{ll*{9}{c}}
\toprule
\textbf{Model} & \textbf{Classifier} &
\makecell{\textbf{ARC-}\\\textbf{Challenge}} &
\makecell{\textbf{ARC-}\\\textbf{Easy}} &
\makecell{\textbf{Commonsense-}\\\textbf{QA}} &
\textbf{MMLU} &
\makecell{\textbf{OpenBook-}\\\textbf{QA}} &
\textbf{HellaSwag} &
\makecell{\textbf{Wino-}\\\textbf{Grande}} &
\textbf{PIQA} &
\makecell{\textbf{SocialI-}\\\textbf{QA}}\\
\midrule

\multicolumn{11}{l}{\emph{Llama\cite{grattafiori2024llama3herdmodels}}} \\
\addlinespace[0.3ex]

\multirow{2}{*}{Llama 3.2 1B} & Self-labeled
& \onescore{46.3}{+2.0} & \onescore{69.9}{+2.5} & \onescore{52.1}{+0.4}
& \onescore{43.8}{+1.5} & \onescore{47.4}{+3.8} & \onescore{71.8}{+9.0}
& \onescore{76.4}{+16.7} & \onescore{88.4}{+11.7} & \onescore{48.5}{+0.5}\\
& Majority-labeled
& \onescore{43.9}{-0.4} & \onescore{67.5}{+0.1} & \onescore{51.6}{-0.1}
& \onescore{41.1}{-1.1} & \onescore{44.0}{+0.4} & \onescore{71.8}{+9.0}
& \onescore{69.8}{+10.1} & \onescore{88.4}{+11.7} & \onescore{48.1}{+0.1}\\
\cmidrule(lr){2-11}
\noalign{\vskip -1.5ex}
\addlinespace

\multirow{2}{*}{Llama 3.2 1B Inst} & Self-labeled
& \onescore{54.9}{+0.3} & \onescore{76.5}{+1.2} & \onescore{59.3}{+0.2}
& \onescore{45.9}{+0.1} & \onescore{59.8}{+0.0} & \onescore{68.9}{+9.1}
& \onescore{68.2}{+10.1} & \onescore{82.0}{+7.8} & \onescore{55.4}{+0.4}\\
& Majority-labeled
& \onescore{54.2}{-0.4} & \onescore{75.2}{-0.1} & \onescore{59.1}{+0.0}
& \onescore{45.6}{-0.2} & \onescore{59.6}{-0.2} & \onescore{68.9}{+9.1}
& \onescore{61.2}{+3.1} & \onescore{80.6}{+6.4} & \onescore{55.1}{+0.1}\\
\cmidrule(lr){2-11}
\noalign{\vskip -1.5ex}
\addlinespace

\multirow{2}{*}{Llama 3.2 3B} & Self-labeled
& \onescore{68.0}{+0.3} & \onescore{84.8}{+0.3} & \onescore{67.6}{+1.1}
& \onescore{55.9}{-1.1} & \onescore{65.4}{+0.6} & \onescore{79.0}{+6.3}
& \onescore{73.0}{+4.7} & \onescore{86.6}{+8.8} & \onescore{59.9}{+0.1}\\
& Majority-labeled
& \onescore{67.7}{+0.0} & \onescore{84.5}{+0.0} & \onescore{66.9}{+0.4}
& \onescore{55.4}{-1.6} & \onescore{64.8}{+0.0} & \onescore{79.0}{+6.3}
& \onescore{60.1}{-8.2} & \onescore{86.4}{+8.6} & \onescore{59.9}{+0.2}\\
\cmidrule(lr){2-11}
\noalign{\vskip -1.5ex}
\addlinespace

\multirow{2}{*}{Llama 3.2 3B Inst} & Self-labeled
& \onescore{74.1}{+0.4} & \onescore{88.3}{+0.3} & \onescore{74.0}{+0.2}
& \onescore{58.5}{-0.1} & \onescore{76.2}{+0.0} & \onescore{72.8}{+2.8}
& \onescore{70.0}{+5.9} & \onescore{79.3}{+2.5} & \onescore{68.0}{+0.3}\\
& Majority-labeled
& \onescore{74.1}{+0.4} & \onescore{88.3}{+0.2} & \onescore{73.9}{+0.0}
& \onescore{58.4}{-0.2} & \onescore{76.4}{+0.2} & \onescore{65.3}{-4.7}
& \onescore{60.9}{-3.2} & \onescore{74.0}{-2.8} & \onescore{68.1}{+0.4}\\
\cmidrule(lr){2-11}
\noalign{\vskip -1.5ex}
\addlinespace

\multirow{2}{*}{Llama 3.1 8B} & Self-labeled
& \onescore{80.1}{+0.7} & \onescore{92.1}{+0.3} & \onescore{71.3}{+0.5}
& \onescore{63.9}{-0.8} & \onescore{76.6}{+1.0} & \onescore{83.6}{+5.1}
& \onescore{78.5}{+8.0} & \onescore{91.0}{+9.7} & \onescore{62.7}{+0.0}\\
& Majority-labeled
& \onescore{79.7}{+0.3} & \onescore{92.0}{+0.1} & \onescore{71.0}{+0.2}
& \onescore{63.6}{-1.1} & \onescore{76.6}{+1.0} & \onescore{83.6}{+5.1}
& \onescore{76.2}{+5.6} & \onescore{90.9}{+9.7} & \onescore{62.6}{-0.1}\\
\cmidrule(lr){2-11}
\noalign{\vskip -1.5ex}
\addlinespace

\multirow{2}{*}{Llama 3.1 8B Inst} & Self-labeled
& \onescore{81.9}{-0.2} & \onescore{93.2}{+0.0} & \onescore{76.1}{+0.2}
& \onescore{65.3}{-0.1} & \onescore{82.4}{+1.4} & \onescore{79.9}{+2.7}
& \onescore{71.0}{+0.3} & \onescore{82.2}{+2.0} & \onescore{70.7}{+0.0}\\
& Majority-labeled
& \onescore{81.8}{-0.3} & \onescore{93.2}{+0.0} & \onescore{75.9}{+0.0}
& \onescore{65.3}{-0.1} & \onescore{81.8}{+0.8} & \onescore{79.7}{+2.5}
& \onescore{63.8}{-6.9} & \onescore{80.1}{-0.2} & \onescore{70.6}{-0.2}\\

\midrule
\multicolumn{11}{l}{\emph{Mistral\cite{jiang2023mistral7b}}} \\
\addlinespace[0.3ex]

\multirow{2}{*}{Mistral 7B v0.3} & Self-labeled
& \onescore{76.8}{+0.2} & \onescore{88.6}{+0.0} & \onescore{59.4}{+1.9}
& \onescore{58.7}{-1.3} & \onescore{68.8}{+0.6} & \onescore{83.0}{+4.5}
& \onescore{78.5}{+6.9} & \onescore{88.8}{+7.6} & \onescore{60.8}{-0.3}\\
& Majority-labeled
& \onescore{76.6}{+0.0} & \onescore{88.4}{-0.2} & \onescore{57.2}{-0.2}
& \onescore{58.5}{-1.5} & \onescore{68.4}{+0.2} & \onescore{82.9}{+4.3}
& \onescore{72.7}{+1.1} & \onescore{88.6}{+7.4} & \onescore{61.1}{-0.1}\\
\cmidrule(lr){2-11}
\noalign{\vskip -1.5ex}
\addlinespace

\multirow{2}{*}{Mistral 7B Inst v0.3} & Self-labeled
& \onescore{78.2}{+0.3} & \onescore{88.5}{+0.3} & \onescore{70.1}{+0.7}
& \onescore{59.3}{-1.0} & \onescore{74.8}{+1.2} & \onescore{78.9}{-2.3}
& \onescore{74.3}{-0.7} & \onescore{81.5}{-1.1} & \onescore{68.9}{+0.3}\\
& Majority-labeled
& \onescore{78.0}{+0.1} & \onescore{88.3}{+0.1} & \onescore{69.6}{+0.2}
& \onescore{58.7}{-1.6} & \onescore{74.2}{+0.6} & \onescore{80.8}{-0.4}
& \onescore{71.9}{-3.1} & \onescore{83.4}{+0.7} & \onescore{68.9}{+0.3}\\
\cmidrule(lr){2-11}
\noalign{\vskip -1.5ex}
\addlinespace

\multirow{2}{*}{Mistral Small 24B Base 2501} & Self-labeled
& \onescore{92.9}{+0.2} & \onescore{97.8}{+0.0} & \onescore{70.9}{+0.1}
& \onescore{76.7}{-1.5} & \onescore{85.6}{+0.4} & \onescore{85.5}{+4.5}
& \onescore{77.6}{+0.9} & \onescore{91.3}{+8.2} & \onescore{70.2}{+0.2}\\
& Majority-labeled
& \onescore{92.9}{+0.2} & \onescore{97.8}{+0.0} & \onescore{70.8}{+0.0}
& \onescore{76.6}{-1.6} & \onescore{85.6}{+0.4} & \onescore{85.4}{+4.4}
& \onescore{74.0}{-2.7} & \onescore{91.3}{+8.1} & \onescore{70.2}{+0.2}\\
\cmidrule(lr){2-11}
\noalign{\vskip -1.5ex}
\addlinespace

\multirow{2}{*}{Mistral Small 24B Inst 2501} & Self-labeled
& \onescore{93.9}{+0.0} & \onescore{98.2}{+0.0} & \onescore{82.6}{+0.4}
& \onescore{78.7}{-0.6} & \onescore{91.6}{-0.2} & \onescore{87.2}{+0.9}
& \onescore{79.7}{+3.5} & \onescore{88.8}{+3.8} & \onescore{79.2}{+0.2}\\
& Majority-labeled
& \onescore{93.9}{+0.0} & \onescore{98.2}{+0.0} & \onescore{82.3}{+0.1}
& \onescore{78.5}{-0.8} & \onescore{91.2}{-0.6} & \onescore{83.1}{-3.2}
& \onescore{72.7}{-3.6} & \onescore{88.3}{+3.3} & \onescore{79.2}{+0.2}\\
\cmidrule(lr){2-11}
\noalign{\vskip -1.5ex}
\addlinespace

\multirow{2}{*}{Mistral Nemo Base 2407} & Self-labeled
& \onescore{81.0}{-0.1} & \onescore{91.0}{-0.1} & \onescore{58.6}{+1.1}
& \onescore{62.6}{-1.3} & \onescore{73.0}{-0.2} & \onescore{84.3}{+4.3}
& \onescore{78.5}{+4.8} & \onescore{89.5}{+7.3} & \onescore{64.7}{-0.2}\\
& Majority-labeled
& \onescore{80.7}{-0.3} & \onescore{91.0}{-0.1} & \onescore{55.9}{-1.6}
& \onescore{62.3}{-1.6} & \onescore{73.4}{+0.2} & \onescore{84.2}{+4.3}
& \onescore{67.7}{-5.9} & \onescore{89.2}{+7.0} & \onescore{64.9}{+0.1}\\
\cmidrule(lr){2-11}
\noalign{\vskip -1.5ex}
\addlinespace

\multirow{2}{*}{Mistral Nemo Inst 2407} & Self-labeled
& \onescore{75.0}{+0.9} & \onescore{86.7}{+2.7} & \onescore{68.5}{+6.6}
& \onescore{60.2}{+1.0} & \onescore{70.0}{+4.2} & \onescore{83.0}{+2.8}
& \onescore{81.7}{+7.2} & \onescore{86.0}{+4.3} & \onescore{73.4}{-0.1}\\
& Majority-labeled
& \onescore{74.0}{-0.1} & \onescore{84.0}{-0.1} & \onescore{61.9}{+0.0}
& \onescore{58.7}{-0.4} & \onescore{66.2}{+0.4} & \onescore{81.0}{+0.7}
& \onescore{54.9}{-19.6} & \onescore{85.4}{+3.7} & \onescore{73.4}{-0.1}\\

\bottomrule
\end{tabular}
\end{adjustbox}

\caption{Model performance of Llama 3.1, 3.2  and Mistral families using the classifier trained on model-generated labels. Values in parentheses indicate the performance gain relative to the baseline with the preferred format ($\max(\text{baseline}_{\texttt{symbol}}, \text{baseline}_{\texttt{cloze}})$).}

\label{tab:model_leveraged_full_llama_mistral}
\end{table}

\begin{table}[!t]
\centering
\footnotesize
\setlength{\tabcolsep}{3pt}
\renewcommand{\arraystretch}{1.05}

\begin{adjustbox}{max width=\textwidth}
\begin{tabular}{ll*{9}{c}}
\toprule
\textbf{Model} & \textbf{Classifier} &
\makecell{\textbf{ARC-}\\\textbf{Challenge}} &
\makecell{\textbf{ARC-}\\\textbf{Easy}} &
\makecell{\textbf{Commonsense-}\\\textbf{QA}} &
\textbf{MMLU} &
\makecell{\textbf{OpenBook-}\\\textbf{QA}} &
\textbf{HellaSwag} &
\makecell{\textbf{Wino-}\\\textbf{Grande}} &
\textbf{PIQA} &
\makecell{\textbf{SocialI-}\\\textbf{QA}}\\
\midrule

\multicolumn{11}{l}{\emph{Qwen3\cite{yang2025qwen3}}} \\
\addlinespace[0.3ex]

\multirow{2}{*}{Qwen3 0.6B Base} & Self-labeled
& \onescore{62.9}{+0.1} & \onescore{81.4}{+0.0} & \onescore{58.6}{+0.0}
& \onescore{49.7}{-1.7} & \onescore{57.8}{-0.2} & \onescore{59.0}{+8.3}
& \onescore{72.2}{+15.7} & \onescore{77.4}{+6.3} & \onescore{57.9}{-0.8}\\
& Majority-labeled
& \onescore{62.8}{+0.0} & \onescore{81.4}{+0.0} & \onescore{58.6}{+0.0}
& \onescore{49.7}{-1.6} & \onescore{57.8}{-0.2} & \onescore{58.8}{+8.1}
& \onescore{40.0}{-16.5} & \onescore{73.4}{+2.3} & \onescore{58.0}{-0.7}\\
\cmidrule(lr){2-11}
\noalign{\vskip -1.5ex}
\addlinespace

\multirow{2}{*}{Qwen3 0.6B} & Self-labeled
& \onescore{52.3}{+0.3} & \onescore{71.9}{+0.3} & \onescore{48.4}{+1.7}
& \onescore{42.0}{-1.0} & \onescore{53.0}{+0.8} & \onescore{58.1}{+13.1}
& \onescore{81.2}{+28.3} & \onescore{82.6}{+15.9} & \onescore{52.0}{-0.3}\\
& Majority-labeled
& \onescore{52.0}{+0.0} & \onescore{71.6}{+0.0} & \onescore{46.7}{+0.0}
& \onescore{41.4}{-1.6} & \onescore{52.2}{+0.0} & \onescore{58.1}{+13.1}
& \onescore{72.7}{+19.8} & \onescore{82.5}{+15.8} & \onescore{52.0}{-0.3}\\
\cmidrule(lr){2-11}
\noalign{\vskip -1.5ex}
\addlinespace

\multirow{2}{*}{Qwen3 1.7B Base} & Self-labeled
& \onescore{80.3}{+0.0} & \onescore{91.9}{+0.1} & \onescore{72.6}{+0.1}
& \onescore{60.4}{-2.5} & \onescore{72.2}{+0.2} & \onescore{71.8}{+8.6}
& \onescore{69.3}{+8.2} & \onescore{82.0}{+7.2} & \onescore{69.4}{+0.4}\\
& Majority-labeled
& \onescore{80.3}{+0.0} & \onescore{91.8}{+0.0} & \onescore{72.4}{-0.1}
& \onescore{60.3}{-2.6} & \onescore{72.0}{+0.0} & \onescore{71.8}{+8.6}
& \onescore{54.2}{-6.9} & \onescore{82.0}{+7.2} & \onescore{69.4}{+0.5}\\
\cmidrule(lr){2-11}
\noalign{\vskip -1.5ex}
\addlinespace

\multirow{2}{*}{Qwen3 1.7B} & Self-labeled
& \onescore{73.9}{+0.0} & \onescore{87.9}{+0.1} & \onescore{64.3}{+0.0}
& \onescore{55.0}{-2.1} & \onescore{65.8}{+0.0} & \onescore{67.9}{+10.3}
& \onescore{66.2}{+8.1} & \onescore{83.8}{+11.2} & \onescore{62.5}{+0.2}\\
& Majority-labeled
& \onescore{73.9}{+0.0} & \onescore{87.8}{+0.0} & \onescore{64.3}{+0.0}
& \onescore{54.9}{-2.2} & \onescore{66.0}{+0.2} & \onescore{67.9}{+10.3}
& \onescore{61.1}{+3.0} & \onescore{83.6}{+11.0} & \onescore{62.5}{+0.2}\\
\cmidrule(lr){2-11}
\noalign{\vskip -1.5ex}
\addlinespace

\multirow{2}{*}{Qwen3 4B Base} & Self-labeled
& \onescore{88.1}{+0.0} & \onescore{96.4}{+0.0} & \onescore{82.5}{+0.0}
& \onescore{70.1}{-2.6} & \onescore{81.4}{+0.0} & \onescore{76.1}{+5.3}
& \onescore{68.8}{+2.8} & \onescore{83.5}{+3.4} & \onescore{74.0}{+0.2}\\
& Majority-labeled
& \onescore{88.1}{+0.0} & \onescore{96.4}{+0.0} & \onescore{82.4}{-0.1}
& \onescore{69.9}{-2.7} & \onescore{81.6}{+0.2} & \onescore{76.4}{+5.7}
& \onescore{67.1}{+1.0} & \onescore{82.2}{+2.1} & \onescore{74.0}{+0.2}\\
\cmidrule(lr){2-11}
\noalign{\vskip -1.5ex}
\addlinespace

\multirow{2}{*}{Qwen3 4B} & Self-labeled
& \onescore{87.1}{+0.0} & \onescore{94.1}{+0.0} & \onescore{75.7}{+0.0}
& \onescore{67.1}{-2.4} & \onescore{78.4}{+0.4} & \onescore{71.0}{+6.2}
& \onescore{65.8}{+4.5} & \onescore{85.4}{+10.3} & \onescore{72.2}{-0.1}\\
& Majority-labeled
& \onescore{87.1}{+0.0} & \onescore{94.1}{+0.0} & \onescore{75.6}{-0.1}
& \onescore{67.0}{-2.5} & \onescore{78.0}{+0.0} & \onescore{70.8}{+5.9}
& \onescore{66.1}{+4.8} & \onescore{85.3}{+10.2} & \onescore{72.2}{-0.1}\\
\cmidrule(lr){2-11}
\noalign{\vskip -1.5ex}
\addlinespace

\multirow{2}{*}{Qwen3 8B Base} & Self-labeled
& \onescore{92.5}{+0.0} & \onescore{97.3}{+0.1} & \onescore{86.2}{+0.3}
& \onescore{73.5}{-2.5} & \onescore{84.2}{-0.8} & \onescore{81.2}{+5.4}
& \onescore{71.3}{+2.8} & \onescore{88.2}{+8.8} & \onescore{78.8}{+0.2}\\
& Majority-labeled
& \onescore{92.5}{+0.0} & \onescore{97.3}{+0.1} & \onescore{86.2}{+0.2}
& \onescore{73.5}{-2.5} & \onescore{84.4}{-0.6} & \onescore{81.2}{+5.4}
& \onescore{68.1}{-0.4} & \onescore{87.8}{+8.3} & \onescore{78.8}{+0.2}\\
\cmidrule(lr){2-11}
\noalign{\vskip -1.5ex}
\addlinespace

\multirow{2}{*}{Qwen3 8B} & Self-labeled
& \onescore{91.2}{+0.0} & \onescore{96.2}{+0.1} & \onescore{78.6}{+0.1}
& \onescore{71.6}{-2.5} & \onescore{85.8}{+0.8} & \onescore{78.5}{+6.9}
& \onescore{72.1}{+6.6} & \onescore{88.8}{+10.3} & \onescore{75.8}{+0.2}\\
& Majority-labeled
& \onescore{91.1}{-0.1} & \onescore{96.2}{+0.1} & \onescore{78.5}{+0.0}
& \onescore{71.4}{-2.7} & \onescore{85.6}{+0.6} & \onescore{78.5}{+6.9}
& \onescore{59.0}{-6.5} & \onescore{88.7}{+10.2} & \onescore{75.8}{+0.2}\\
\cmidrule(lr){2-11}
\noalign{\vskip -1.5ex}
\addlinespace

\multirow{2}{*}{Qwen3 14B Base} & Self-labeled
& \onescore{93.6}{+0.0} & \onescore{98.0}{+0.0} & \onescore{87.0}{+0.2}
& \onescore{77.0}{-2.5} & \onescore{90.8}{+0.4} & \onescore{83.8}{+4.9}
& \onescore{74.3}{+4.1} & \onescore{88.1}{+7.5} & \onescore{79.9}{+0.0}\\
& Majority-labeled
& \onescore{93.6}{+0.0} & \onescore{98.0}{+0.0} & \onescore{86.9}{+0.2}
& \onescore{76.9}{-2.6} & \onescore{90.8}{+0.4} & \onescore{83.8}{+5.0}
& \onescore{74.7}{+4.5} & \onescore{89.3}{+8.7} & \onescore{79.8}{-0.1}\\
\cmidrule(lr){2-11}
\noalign{\vskip -1.5ex}
\addlinespace

\multirow{2}{*}{Qwen3 14B} & Self-labeled
& \onescore{92.4}{+0.1} & \onescore{97.5}{+0.0} & \onescore{80.0}{-0.1}
& \onescore{76.5}{-1.6} & \onescore{90.4}{+0.4} & \onescore{82.2}{+5.9}
& \onescore{75.4}{+4.0} & \onescore{89.7}{+9.4} & \onescore{75.9}{+0.1}\\
& Majority-labeled
& \onescore{92.4}{+0.1} & \onescore{97.4}{+0.0} & \onescore{79.9}{-0.2}
& \onescore{76.2}{-1.9} & \onescore{90.0}{+0.0} & \onescore{82.2}{+5.9}
& \onescore{73.1}{+1.7} & \onescore{89.6}{+9.3} & \onescore{75.8}{+0.1}\\

\bottomrule
\end{tabular}
\end{adjustbox}

\caption{Model performance of Qwen 3 family using the classifier trained on model-generated labels. Values in parentheses indicate the performance gain relative to the baseline with the preferred format ($\max(\text{baseline}_{\texttt{symbol}}, \text{baseline}_{\texttt{cloze}})$).}
\label{tab:model_leveraged_full_qwen}
\end{table}
%\FloatBarrier

\end{document}